%% file: main.tex
\documentclass{article}
\usepackage{iclr2027_conference,times}
\iclrfinalcopy
\usepackage[utf8]{inputenc}
\usepackage[T1]{fontenc}
\usepackage{microtype,graphicx,subcaption,booktabs,multirow}
\usepackage{amsmath,amssymb,amsfonts,mathtools,amsthm}
\usepackage{hyperref,url}
\hypersetup{
  hidelinks,
  pdftitle={Simple Diffusion Language Models Are More Effective Few-Step Generators Than Reported},
  pdfauthor={Hasan Amin, Ming Yin, Rajiv Khanna}
}
\usepackage[capitalize,noabbrev]{cleveref}
\theoremstyle{plain}
\newtheorem{theorem}{Theorem}
\newtheorem{proposition}{Proposition}

\theoremstyle{remark}

\crefname{theorem}{Theorem}{Theorems}
\crefname{proposition}{Proposition}{Propositions}
\crefname{lemma}{Lemma}{Lemmas}
\crefname{corollary}{Corollary}{Corollaries}
\crefname{equation}{}{}
\DeclareMathOperator{\TV}{TV}
\DeclareMathOperator{\TC}{TC}
\DeclareMathOperator{\Var}{Var}
\newcommand{\KL}[2]{\operatorname{KL}\!\left(#1\,\middle\|\,#2\right)}
\newcommand{\E}{\mathbb E}
\newcommand{\J}{\mathcal J}
\newcommand{\Bern}{\operatorname{Bern}}
\newcommand{\hbin}{H_{\mathrm b}}
\newcommand{\logit}{\operatorname{logit}}
\newcommand{\sharpm}{MDLM$^{\sharp}$}
\title{Simple Diffusion Language Models Are More Effective Few-Step Generators Than Reported}
\author{Hasan Amin, Ming Yin, Rajiv Khanna \\
Purdue University \\
\texttt{\{hasanamin,mingyin,rajivak\}@purdue.edu}
}
\begin{document}
\maketitle
% The ICLR style sets its publication header in \maketitle; clear it here.
\fancyhead{}
\renewcommand{\headrulewidth}{0pt}

\begin{abstract}
    Diffusion language models (DLMs) promise fast parallel generation, yet high-quality samples often require large number of refinement steps, which diminishes their advantage in practice. This has led to massive interest in and rapid development of new methods for effective few-step generation. 
    We show that much of the supposed quality gap at few steps can instead arise from a suboptimally configured sampler. Modest sampler sharpening, without any model retraining, enables a couple years old masked DLM to rival supposedly far improved successors. This differently sampled DLM in fact achieves lower generative perplexity in just $16$ steps than what its standard sampler obtains with $1024$, while improving both judged quality and semantic diversity. We further show that conventional per-output metrics can fundamentally obscure these gains, since any optimal trade-off between two such metrics can be attained by a generator supported on at most two outputs. We subsequently introduce \textsc{GroupEval}, which separately evaluates quality and across-output semantic diversity, and offers fresh insights including uncovering how $1.5-4.7\times$ perplexity gains of a distilled model yield no corresponding quality gain. Finally, we explain why sharpening helps: parallel unmasking destroys dependencies among simultaneously generated tokens, creating a gap between prediction and generation. We prove that pervasive temperature choice of one is generically suboptimal under parallel sampling even for an exact denoiser, and that worse predictions can yield better samples. Through these results, we argue for a broader evaluation principle of treating the deployed generator as the object of comparison, benchmarking it against tuned baselines, and assessing quality and diversity jointly and with more human-aligned measures.
\end{abstract}

\input{sections/intro}
\input{sections/audit}
\input{sections/groupeval}
\input{sections/theory}
\input{sections/conclusion}
\clearpage
\section*{Reproducibility statement}
Detailed proofs appear in \Cref{app:theory,app:collapse_proof}. All relevant code, including precise prompts, is available at: \href{https://github.com/shasanamin/more-effective-dlms}{https://github.com/shasanamin/more-effective-dlms}.
% \Cref{app:setup,app:full_grids,app:certificate_measurement} specify generation and held-out evaluation. \Cref{app:ge_default} fixes the default GroupEval configuration, and the GroupEval package provides frozen rubrics, grouping and scoring rules, evaluator manifests, aggregate scores, and uncertainty procedures. All figures are rebuilt from portable aggregates without model inference. Raw generations and individual judgments are not redistributed, so exact reaggregation and inspection of semantic assignments require the original archives. Repeating generation requires the named checkpoints and evaluation environment; repeating the headline comparisons requires no model training.

\section*{Ethics statement}
This work evaluates text generation and does not involve a new human-subject study. Web-trained generators may produce harmful, biased, or false text. GroupEval measures rubric-based perceived quality and local semantic diversity, not factuality, safety, or demographic fairness. Judge biases and changes to hosted models can affect its scores. Our protocol preserves judge-specific results and records provenance rather than treating automated judgments as human ground truth.
\input{ai_use_statement}
\bibliographystyle{iclr2027_conference}
\bibliography{refs}
\appendix
\clearpage
\input{sections/app_exp}
% \clearpage
\input{sections/group_eval_appendix}
% \clearpage
\input{sections/app_theory}
% \clearpage
\input{sections/app_samplers}
\end{document}

%% file: sections/intro.tex
\section{Introduction}
\label{sec:intro}
Diffusion language models promise fast text generation by denoising many tokens in parallel, yet their sample quality falters at few steps. Efforts to close this gap span distillation~\citep{deschenaux2024sdtt,hayakawa2025di4c,sahoo2025duo,zheng2025didi}, new diffusion processes~\citep{lou2024sedd,vonrutte2025gidd,pynadath2025candi}, and new samplers~\citep{zheng2024masked,liu2025ddpd,kim2025train}. However, there's a critical nuance that is often neglected: training evaluates a denoiser, while deployment runs a finite-step generator built from its predictions. Better predictions may not necessarily produce better samples at a fixed sampling budget. We study this gap in the original masked diffusion language model (MDLM)~\citep{austin2021d3pm,sahoo2024simple,shi2024simplified,ou2024radd}. Its standard sampler uses temperature one and 64-bit floating-point noise (fp64), adopted after \citet{zheng2024masked} showed that 32-bit (fp32) categorical sampling silently sharpens draws and made earlier comparisons unfair.

\begin{figure}[t]
\centering
\includegraphics[width=\linewidth]{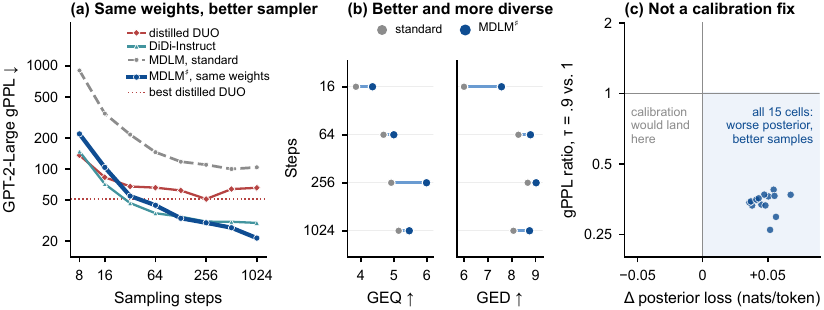}
\caption{\textbf{Same weights, sharper sampler result in a better generator from a worse predictor.} \textbf{(a)} The unchanged public MDLM, sampled as \sharpm{} $(\tau{=}.9,\mathrm{fp32})$ instead of the standard $(\tau{=}1,\mathrm{fp64})$, rivals distilled successors evaluated under one protocol. \textbf{(b)} Under the default \textsc{GroupEval} judge, quality and semantic diversity \emph{both} rise despite fall in token entropy. \textbf{(c)} Across three checkpoints and five budgets, sharpening to $\tau=.9$ worsens held-out posterior cross-entropy in all 15 cases (paired 95\% intervals above zero) while cutting gPPL $2.6$--$3.8\times$. 
\Cref{sec:theory} explains the prediction--generation gap.
}
\label{fig:overview}
\vspace{-6pt}
\end{figure}

We find that much of the few-step gap lies in the sampler. Holding the well-trained, public MDLM checkpoint fixed and sampling it slightly sharper, at temperature $.9$ and in fp32 (\sharpm), gives lower generative perplexity (gPPL, perplexity under an external pretrained model) at 64 steps than the standard sampler reaches at 1,024: $41$ versus $107$, and $53$ with temperature alone. Evaluated identically to its successors, \sharpm{} at 64 steps beats the best result of distilled DUO~\citep{sahoo2025duo} at \emph{any} budget, and from 128 steps on it matches DiDi-Instruct~\citep{zheng2025didi} (\Cref{fig:overview}a). The standard sampler can also reverse research conclusions at the small budgets where parallel generation is most valuable. A checkpoint trained $44\times$ longer, or a model with $5\times$ more parameters, becomes a \emph{worse} 2--8-step generator based on gPPL, even though the longer-trained checkpoint has better held-out posterior loss throughout (\Cref{sec:less}).

Temperature is not a new knob knob~\citep{holtzman2020nucleus,chang2022maskgit}, and it has even been observed recently how it can substitute for steps along MDLM's perplexity--entropy frontier \citep{pynadath2025candi}. But does a better point on that frontier mean a better \emph{generator}? The standard metrics cannot tell. Perplexity and token entropy grade each output on its own, so neither notices when a generator keeps producing the same good output. And this blind spot is structural. We prove that every optimal trade-off between two per-output scores can be reached by a generator with at most two distinct outputs (\Cref{prop:collapse}). A stronger evaluator model does not help, because the problem lies in scoring outputs one at a time. What is needed is an evaluation that compares outputs with each other, and \textsc{GroupEval} provides it. A blinded judge reads a group of outputs from one generator, scores each for quality, and sorts them by meaningfully distinct content, yielding \textsc{GroupEval Quality (GEQ)} and \textsc{GroupEval Diversity (GED)}. \textsc{GEQ} asks whether each output is good. \textsc{GED} asks whether the outputs say different things. With both scores, \sharpm{} is better \emph{and} more diverse across sampling budgets (\Cref{fig:overview}b). On matched outputs, two likelihood evaluators prefer DiDi-Instruct to MDLM by $1.5$--$4.7\times$ in perplexity, while \textsc{GroupEval} favors MDLM in quality and diversity. We find DiDi's samples to repeat fluent passages, a pattern that per-output scores need not penalize.

One may wonder why should a trained denoiser needs sharpening. Sharpening in fact \emph{worsens} held-out posterior loss at every checkpoint and budget measured, even as samples improve (\Cref{fig:overview}c). In parallel unmasking, tokens revealed together are drawn independently, so their dependence is lost even when every prediction is exact. Asked to fill in a two-token city name in one round, the sampler can pair ``New'' with ``Angeles''. Our finite-step analysis makes this precise. It splits the sampler's error exactly into the model's prediction error, the dependence lost by drawing tokens together, and a credit from averaging over reveal orders that changes with temperature (\Cref{thm:accounting}). Because that credit moves, the temperature that is best for prediction is generally not the one that is best for generation. Even for an exact denoiser, temperature one is almost never optimal once tokens are revealed in parallel (\Cref{thm:generic}), and in a two-token example sharpening makes the sampler exact while worsening its predictions (\Cref{prop:repair}). Better predictors therefore need not be better few-step generators, and held-out loss alone cannot certify a sampler.

Our key contributions are as follows:
\begin{enumerate}
\setlength\itemsep{1pt}
\item \textbf{A stronger baseline that changes comparisons.} Tuning only temperature and categorical precision lets an unchanged checkpoint by \citet{sahoo2024simple} rival its distilled successors, and it reverses apparent few-step benefits of longer training and larger models (\Cref{sec:audit}).
\item \textbf{A reusable group-level evaluation.} We prove that per-output metrics cannot detect collapse and introduce \textsc{GEQ} and \textsc{GED}, two separately reported scores under a fixed protocol. They confirm the sharpening gain and find no quality lead behind a $1.5$--$4.7\times$ perplexity lead (\Cref{sec:groupeval}).
\item \textbf{A theory of the prediction--generation gap.} An exact finite-step accounting of parallel unmasking shows that temperature one is generically suboptimal even for an exact denoiser, and that a worse predictor can be the better few-step generator (\Cref{sec:theory}).
\end{enumerate}

\paragraph{Relation to prior work.}
Temperature--step trade-offs in DLMs have been explored~\citep{pynadath2025candi,shah2026care}, and prior theory quantifies dependence lost in parallel draws~\citep{lavenant2025error,chen2025optimal}. Here the tuned baseline changes model rankings, while our accounting includes learned, tempered denoisers and path mixing. \textsc{GroupEval} builds on semantic diversity and LLM judging~\citep{kuhn2023semantic,zhang2025noveltybench,zheng2023judging} by fixing a group-level protocol for unconditional generation and testing where conventional metrics disagree. \Cref{app:samplers} develops these connections.

%% file: sections/audit.tex
\section{The Few-Step Gap Is Largely a Sampling Artifact}
\label{sec:setup}\label{sec:experiments}\label{sec:audit}
\paragraph{A generator is a checkpoint plus a sampler.}
A few-step MDLM generator is indexed by $(\theta,M,\tau,\kappa)$: checkpoint, sampling steps, temperature, and categorical precision. From an all-mask sequence, each of the $M$ ancestral updates independently decides which masked positions to reveal and draws their tokens from the denoiser's tempered clean-token distribution
\begin{equation}
q_{\theta,\lambda}(v\mid z)=\frac{\exp\{\lambda\, r_\theta(v\mid z)\}}{\sum_u\exp\{\lambda\, r_\theta(u\mid z)\}},\qquad \lambda=1/\tau,
\label{eq:temperature_calibration}
\end{equation}
where $r_\theta$ are logits given the partially masked sequence $z$ (revealed tokens are kept). The precision $\kappa$ is that of the random numbers behind each Gumbel-max token draw, not of the network. In fp32 these numbers are too coarse to reach the extreme tail, so rare tokens are drawn too seldom. \citet{zheng2024masked} showed that this lowers the effective temperature and recommended fp64, which became the default in later codebases. Because MDLM draws the mask-or-token outcome in the same categorical, precision also shifts when tokens are revealed, so fp32 is a distinct sharpening, not a relabeled temperature. 
Steps count configured updates $M$, which bound but do not measure network calls (\Cref{app:setup}).

\paragraph{One checkpoint, one sharpening, frozen before any judging.}
We use the unchanged public MDLM-Small checkpoint~\citep{sahoo2024simple}, unconditional OpenWebText generation of 1,024 tokens, and 128 samples per cell, scored by GPT-2 gPPL and by the mean per-sample unigram (token-frequency) entropy $H$ in nats. The \emph{standard} sampler is $(\tau{=}1,\mathrm{fp64})$. \sharpm{} is $(\tau{=}.9,\mathrm{fp32})$, chosen once from a sweep over $\tau\in\{.8,\dots,1.1\}\times\{\mathrm{fp32},\mathrm{fp64}\}$ as the sharpest setting that keeps $H\ge5$ at most budgets (a collapse screen) before any \textsc{GroupEval} judgment, and then frozen for every budget, judge, and model. Weights, schedule, and algorithm never change.

\begin{table}[b]
\centering\small
\caption{\textbf{Two sampling knobs compound: at 64 steps, \sharpm{} decisively outperforms the standard sampler at 1,024 (bold).} GPT-2 gPPL (lower is better) with unigram entropy $H$ in brackets, 128 samples per cell. Temperature and precision each help on their own.}
\label{tab:headline_main}
\setlength{\tabcolsep}{5pt}
\begin{tabular}{rcccc}
\toprule
Steps $M$ & Standard $(1,\mathrm{fp64})$ & $(1,\mathrm{fp32})$ & $(.9,\mathrm{fp64})$ & \sharpm{} $(.9,\mathrm{fp32})$ \\
\midrule
16 & 357.6 [5.79] & 338.7 [5.78] & 109.8 [5.37] & 102.2 [5.35] \\
64 & 154.9 [5.71] & 107.6 [5.62] & 53.1 [5.36] & \textbf{41.3} [5.27] \\
256 & 115.5 [5.66] & 61.5 [5.45] & 40.9 [5.31] & 28.1 [5.13] \\
1024 & \textbf{107.1} [5.63] & 42.5 [5.33] & 36.6 [5.26] & 21.8 [5.01] \\
\bottomrule
\end{tabular}
\end{table}

\paragraph{Temperature beats a $16\times$ step budget, and precision compounds it.}
Temperature alone moves the baseline further than a $16\times$ larger step budget: at fixed fp64, 64 steps at $\tau=.9$ reach gPPL $53.1$, against $107.1$ for the standard sampler at 1,024 (\Cref{tab:headline_main}). Returning to fp32 compounds the gain, to $41.3$ at 64 steps and a $4.9\times$ reduction at 1,024. The ordering is the same for two further MDLM-Small checkpoints from our own training run and for MDLM-Tiny (\Cref{app:baseline_grids,app:tiny,app:full_grids}). The fp32 draw is also cheaper: at 1,024 steps, \sharpm{} uses 40\% less peak memory and runs 17\% faster than the standard sampler (\Cref{app:sampling_cost}).

\paragraph{An unchanged baseline rivals its distilled successors.}
\Cref{fig:overview}a places \sharpm{} among its successors under one protocol: GPT-2-Large gPPL on 40 samples per point, every other model at the standard sampler (\Cref{tab:final_model_compare}). At 64 steps, \sharpm{} scores $44.5$, below the best result of distilled DUO at any budget ($51.2$) and the best of DUO ($70.1$) and SEDD ($100.1$). DiDi-Instruct keeps a perplexity lead through 64 steps. From 128 steps on, \sharpm{} matches or beats it. Sharpening roughly halves DiDi's perplexity as well (\Cref{tab:final_model_compare}), yet \Cref{sec:groupeval} finds that DiDi's perplexity lead is not a quality lead, and that its own sharpening brings no judge-robust quality gain.

\subsection{Less is more, until the sampler is fixed}
\label{sec:less}
\begin{figure}[t]
\centering
\includegraphics[width=\linewidth]{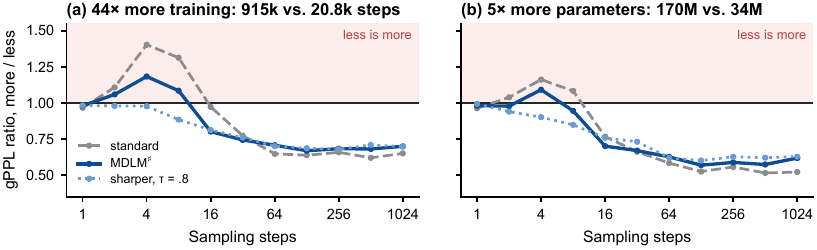}
\caption{\textbf{At the standard sampler, more training and more parameters make few-step generation worse.} Ratio of gPPL for the larger or longer-trained model to the smaller or shorter-trained one, at matched steps and sampler. Values above one (shaded) mean that less is more. \textbf{(a)} MDLM-Small after 915k versus 20.8k training steps. \textbf{(b)} MDLM-Small versus MDLM-Tiny. Less wins only at 2--8 steps, and these inversions shrink under the sharpened sampler $(.9,\mathrm{fp32})$ and vanish with stronger sharpening. GPT-2 gPPL; 128 (Small) and 256 (Tiny) samples per cell.}
\label{fig:less}
\end{figure}
The standard sampler does not merely understate the baseline. It inverts comparisons that guide research decisions. We trained MDLM-Small with the reference recipe and kept checkpoints after 20.8k and 915k steps. At the standard sampler, the longer-trained checkpoint is the worse 2--8-step generator, by $1.40\times$ in gPPL at 4 steps (\Cref{fig:less}a), even though its held-out posterior loss is lower at every audited budget, by $.43$--$.62$ nats per token (\Cref{app:certificate_measurement}). Model size shows the same inversion: the 170M-parameter MDLM-Small is a worse 2--8-step generator than a 34M-parameter MDLM-Tiny (\Cref{fig:less}b). Both inversions shrink under the sharpened sampler and disappear with slightly stronger sharpening ($\tau=.8$). They are also confined to few steps: from 64 steps on, the larger or longer-trained model is $1.4$--$1.9\times$ better in gPPL under all three samplers. A practitioner tuning few-step generation at the standard sampler would stop training early and choose the smaller model. \Cref{prop:repair} shows that such inversions need no optimization pathology: at a fixed temperature, a less accurate predictor can be the better few-step generator.

%% file: sections/groupeval.tex
\section{Are the Gains Real? Evaluate Groups, Not Samples}
\label{sec:groupeval}
Lower perplexity with lower token entropy (\Cref{tab:headline_main}) could also arise from a generator that repeats itself. We need to compare outputs to tell whether sharpening improves generation.
\subsection{Per-sample metrics reward collapse}
\label{sec:collapse}
Generative perplexity and token entropy each grade one sample at a time and average the grades, so copying the best-graded output never costs anything. Formally, for length-$L$ outputs under token-pooled scoring, $\log\mathrm{gPPL}(Q)=\E_{X\sim Q}[-\log R(X)]/L$ for an evaluator $R$, and the reported entropy is $H(Q)=\E_{X\sim Q}\hat H(X)$, the mean unigram entropy of individual outputs. Both are linear in the generator $Q$, and linearity alone makes any protocol built from them degenerate.

\begin{proposition}[Per-sample metrics reward collapse]
\label{prop:collapse}
Let $f$ and $g$ be any per-output scores on a finite output space. Every Pareto-optimal value of $(\E_Qf,\E_Qg)$ over generators $Q$ (in particular, every solution of $\min_Q\E_Qf$ subject to $\E_Qg\ge h$) is attained by a generator supported on at most two outputs.
\end{proposition}
The achievable pairs form the convex hull of per-output points, whose Pareto frontier consists of vertices and edges (\Cref{app:collapse_proof}). Entropy-constrained perplexity minimization, the implicit target of perplexity--entropy reporting~\citep{zheng2024masked,pynadath2025candi}, is therefore solved by repeating two fluent, lexically varied documents. A group of eight samples from such a generator holds at most two distinct outputs, which caps its \textsc{GED} (below) at $3.3$ of $10$. The MDLM samplers we study score $6$--$9$. \textsc{GroupEval} therefore asks directly whether a set of outputs contains good and genuinely different content.

\subsection{GroupEval: quality per output, modes per group}
\label{sec:ge_protocol}
A judge sees a randomly ordered group of $n$ outputs from one generator, with the generator, sampler, budget, and automatic scores hidden. Under a fixed rubric, it scores each output for readability, coherence, substance, and non-degeneration on $\{0,1,2\}$, after a gate that zeroes empty, corrupted, or non-communicative text. It also partitions valid outputs into semantic--discourse modes. Changes of topic, purpose, genre, or discourse structure create modes; paraphrases, entity swaps within a template, formatting, and corruption do not. Deterministic code converts these decisions into two complementary scores. \textsc{\textbf{GroupEval Quality (GEQ)}} is the mean rubric total over \emph{all} outputs, rescaled to $[0,10]$, so failures count. \textsc{\textbf{GroupEval Diversity (\textsc{GED})}} is validity-weighted, normalized mode entropy. For a group $g$ of $n$ outputs with $V_g$ valid outputs and mode shares $p_{gk}$,
\begin{equation}
\mathrm{\textsc{GED}}_g=10\,\frac{V_g}{n}\,
\frac{-\sum_k p_{gk}\log p_{gk}}{\log n},
\qquad
\mathrm{\textsc{GED}}=\frac{1}{|\mathcal G|}\sum_{g\in\mathcal G}\mathrm{\textsc{GED}}_g,
\label{eq:ged_main}
\end{equation}
with $\mathrm{\textsc{GED}}_g=0$ when fewer than two valid outputs or modes remain. Entropy rewards the number and balance of modes, and the validity factor charges for unusable text. With eight outputs (letters are modes, $\times$ is invalid text), the patterns AAAAAAAA, AAAAAAAB, AAAABBBB, and ABCDEFGH score $0$, $1.81$, $3.33$, and $10$; ABCD$\times\times\times\times$ scores $3.33$. Merging modes or invalidating an output never raises \textsc{GED} (\Cref{app:ged_properties}). Valid but weak text keeps its mode mass, which is why \textsc{GEQ} accompanies \textsc{GED}. Read together, they separate the failure modes a single score would merge. Low \textsc{GED} with high \textsc{GEQ} flags good but similar outputs, and high \textsc{GED} with low \textsc{GEQ} flags varied but weak text.

\paragraph{A shared evaluation protocol.}
For unconditional generation we use groups of eight, the open-weight \texttt{gpt-oss-120b} judge, the fixed web-text rubric above, and two fixed partitions of each output pool. Every output is judged twice. \textsc{GEQ} intervals use a partition-blocked Bayesian bootstrap~\citep{rubin1981bayesian}. This blinded, automated procedure lets researchers compare checkpoints, samplers, and budgets under the same rubric without recruiting a new human panel for each comparison. We also conduct exploratory studies with four other judges to test robustness, and a task-specific conditional \textsc{GEQ} check for a separate rubric. These, alongside additional experimental details, are shared in Appendix \ref{app:groupeval}.

\begin{table}[t]
\centering\small
\caption{\textbf{\textsc{GEQ} and \textsc{GED} reveal what gPPL and token entropy miss.} Default judge; groups of eight. \textbf{(A)} Public MDLM, standard $\to$ \sharpm{}, 128 outputs per cell; GPT-2 gPPL. \textbf{(B)} MDLM minus DiDi-Instruct in a separate, matched, interleaved study, 40 outputs per model and cell; the same saved outputs receive \textsc{GroupEval} and GPT-2-Large scores. gPPL ratios above one favor DiDi. 
% $^\ast$Conditional 95\% \textsc{GEQ} interval excludes zero.
}
\label{tab:groupeval_main}\label{tab:judge_differences}\label{tab:crossfamily_main}
\setlength{\tabcolsep}{6pt}
\begin{tabular}{rcccc}
\multicolumn{5}{l}{\textbf{(A) Public MDLM-Small: standard $\to$ \sharpm}}\\
\toprule
Steps & gPPL $\downarrow$ & Token entropy $H$ & \textsc{GEQ} $\uparrow$ & \textsc{GED} $\uparrow$ \\
\midrule
16 & $357.6\to102.2$ & $5.79\to5.35$ & $3.85\to4.36$ & $6.00\to7.57$ \\
64 & $154.9\to\phantom{0}41.3$ & $5.71\to5.27$ & $4.69\to5.00$ & $8.28\to8.79$ \\
256 & $115.5\to\phantom{0}28.1$ & $5.66\to5.13$ & $4.92\to5.99$ & $8.67\to9.02$ \\
1024 & $107.1\to\phantom{0}21.8$ & $5.63\to5.01$ & $5.15\to5.47$ & $8.07\to8.75$ \\
\bottomrule
\end{tabular}

\vspace{6pt}
\setlength{\tabcolsep}{5.2pt}
\begin{tabular}{rccccccc}
\multicolumn{8}{l}{\textbf{(B) MDLM vs.\ DiDi-Instruct on matched outputs}}\\
\toprule
& \multicolumn{3}{c}{Standard sampler} && \multicolumn{3}{c}{Sharpened sampler} \\
\cmidrule(lr){2-4}\cmidrule(lr){6-8}
Steps & gPPL ratio & $\Delta$\textsc{GEQ} & $\Delta$\textsc{GED} && gPPL ratio & $\Delta$\textsc{GEQ} & $\Delta$\textsc{GED} \\
\midrule
16 & $4.58\times$ & $\phantom{+}0.00$ & $+2.61$ && $2.33\times$ & $+1.19$ & $+4.65$ \\
64 & $3.78\times$ & $+0.20$ & $+2.76$ && $2.19\times$ & $+0.41$ & $+1.79$ \\
256 & $3.49\times$ & $+1.50$ & $+2.11$ && $1.76\times$ & $+0.72$ & $+3.04$ \\
1024 & $3.31\times$ & $+0.77$ & $+2.61$ && $1.48\times$ & $+1.06$ & $+2.22$ \\
\bottomrule
\end{tabular}
\end{table}

\subsection{Sharper sampling improves quality and diversity together}
\label{sec:ge_results}
Under the default judge, \sharpm{} raises \textsc{GEQ} and \textsc{GED} at every budget while token entropy falls by $.44$--$.62$ nats (\Cref{tab:groupeval_main}A). Lower lexical entropy accompanies more distinct content. Both factors of \textsc{GED} improve at every budget: at 16 steps, the invalid share falls from $13.3\%$ to $6.6\%$ of outputs, and the effective number of modes among valid outputs per group of eight rises from $4.73$ to $5.74$. The quality gain is judge-robust. All five judges give \sharpm{} the higher \textsc{GEQ} at all four budgets, 13 of these 20 conditional intervals exclude zero, and 16 of the 20 judge--budget pairs also show higher \textsc{GED}. The step savings survive judgment: \sharpm{} at 256 steps beats the standard sampler at 1,024 on \textsc{GEQ} under every judge. Token entropy alone would have rejected this better generator.

\subsection{A perplexity lead is not necessarily a quality lead}
\label{sec:crossfamily}
To compare model families on equal terms, we judge the same saved MDLM and DiDi-Instruct outputs behind \Cref{tab:final_model_compare} in a separate, matched, interleaved study: 40 outputs per model and cell. We rescore those outputs with two likelihood evaluators, GPT-2-Large and Qwen3-8B-Base~\citep{yang2025qwen3}, on the same token spans. The evaluators agree with each other (Spearman $\rho=.98$ across cells) and favor DiDi in all 8 comparisons, by $1.5$--$4.7\times$. \textsc{GroupEval}'s default judge disagrees (\Cref{tab:groupeval_main}B): MDLM has the higher \textsc{GEQ} in 7 comparisons, ties in the eighth, and has the higher \textsc{GED} in all 8, often by a wide margin ($6.6$ versus $2.0$ at 16 steps under the sharpened sampler). The one tie is telling: at 16 steps under the standard sampler, a $4.6\times$ perplexity advantage buys DiDi no judged quality and costs it $2.6$ points of \textsc{GED}. DiDi's outputs often repeat chunks of sensible sentences. A repeated chunk is fluent and predictable, which can lower perplexity, and it can still keep token entropy high. This qualitative observation suggests why the per-output metrics disagree with \textsc{GroupEval}, while the measured disagreement survives first-document scoring and bits-per-byte (\Cref{app:ge_exact}).

\paragraph{\textsc{GroupEval} distinguishes interventions and tracks task success.}
\textsc{GroupEval} does not endorse every perplexity gain. Sharpening DiDi-Instruct roughly halves its gPPL but yields no judge-robust quality gain: every default-judge interval includes zero, and one robustness judge prefers the standard sampler (\Cref{tab:groupeval-didi-all}). With a separate task rubric and no access to answers or tests, conditional \textsc{GEQ} separates correct from incorrect LLaDA-8B and Dream-7B solutions with AUROC $.91$ on GSM8K and $.96$ on HumanEval, and follows the task-dependent ordering of the two models (\Cref{tab:groupeval-conditional-large}).

%% file: sections/theory.tex
\section{Why a Perfect Denoiser Needs Sharpening}
\label{sec:theory}\label{sec:calibration}
\begin{figure}[t]
\centering
\includegraphics[width=\linewidth]{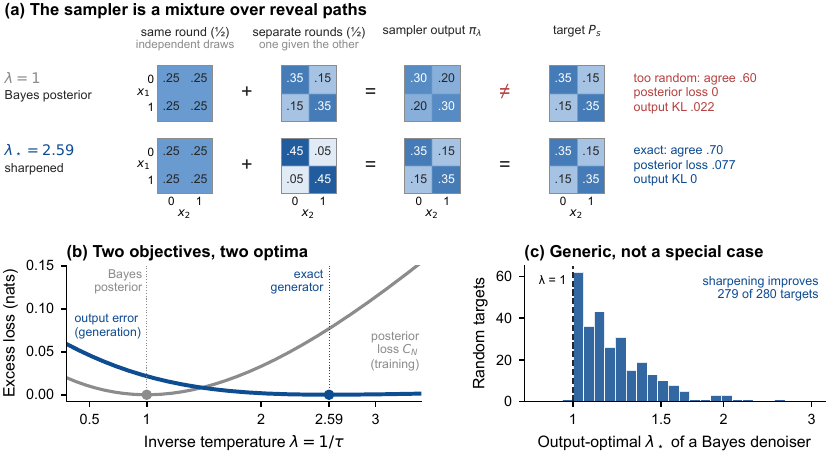}
\caption{\textbf{Training and generation want different temperatures, even for a perfect model.} \textbf{(a)} The sampler as a mixture over reveal paths, for the pair family~\eqref{eq:binary_target} with $s=.7$ and two equal rounds. Cells give the joint law of $(x_1,x_2)$. At $\lambda=1$, the independent same-round branch dilutes the exact sequential branch and the output is too random. At $\lambda_\star\approx2.59$, the sequential branch over-agrees just enough to offset it. \textbf{(b)} The Bayes posterior minimizes posterior loss, but output error vanishes only at $\lambda_\star$. \textbf{(c)} Best inverse temperature for the \emph{exact} Bayes denoiser on 280 random full-support targets ($L\in\{2,3\}$, $|\mathcal V|\le4$, random grids), by enumeration: never $1$, and above $1$ in all but one case.}
\label{fig:theory}\label{fig:bayes_compensation}
\end{figure}

\Cref{sec:audit,sec:groupeval} establish that sharpening yields a better generator. \Cref{fig:overview}c adds a puzzle: the same sharpening worsens posterior prediction loss. Parallel sampling can produce this gap even from exact posteriors. The few-step generator is a \emph{mixture over reveal paths}, each component a product of single-token predictions. Training shapes the components, while generation is judged on the mixture. \Cref{fig:theory}a draws it for two correlated bits and two rounds: half of the paths reveal both bits together and draw them independently, the other half reveal them in turn, and the output law averages the two branches.

\subsection{Training integrates, sampling sums}
\label{sec:entropy_profile}
Let $X\sim P$ on a finite space $\mathcal V^L$. A denoiser predicts a masked coordinate $i$ from the visible values $x_A$ and their positions, $q^i(\cdot\mid x_A)$, and $q^i_\lambda\propto(q^i)^\lambda$ is its tempered version. Its posterior-loss profile and the Bayes entropy profile are
\begin{equation}
 c_\lambda(a)=\sum_{i=1}^L\E_{A,X}[-\log q_\lambda^i(X_i\mid X_A)],
 \qquad
 h(a)=\sum_{i=1}^L\E_A H(X_i\mid X_A),
 \label{eq:profiles}
\end{equation}
where $A\subseteq[L]\setminus\{i\}$ retains each other coordinate independently with probability $a$. A sampler with grid $0=a_0<\cdots<a_N=1$ gives each coordinate an independent reveal round $K_i$ with $\Pr(K_i=k)=\delta_k:=a_k-a_{k-1}$, which is exactly the law of MDLM's ancestral sampler (\Cref{app:reveal_law}). In round $k$, the coordinates $S_k=\{i:K_i=k\}$ are drawn independently from $q_\lambda$ given all earlier rounds. Let $\nu$ be the law of the reveal path $S=(S_1,\ldots,S_N)$, $Q_\lambda(x,S)$ the joint law of output and path, and $\pi_\lambda$ the output law. Predictions are positive and arithmetic is exact. Intuitively, training scores each token against contexts at every masking level, an area under the loss profile, whereas the sampler visits only $N$ levels, a Riemann sum, and never lets same-round tokens see each other.

\begin{theorem}[Training integrates, sampling sums]
\label{thm:accounting}\label{thm:decomposition}
The continuous-time MDLM objective and the data entropy are areas under the two profiles, $\mathcal L_\infty(\lambda)=\int_0^1c_\lambda(a)\,da$ and $H(X)=\int_0^1h(a)\,da$. The few-step sampler evaluates their left Riemann sums instead. With $\J_N(\lambda)=\sum_k\delta_k c_\lambda(a_{k-1})$,
\begin{align}
 \KL{P\otimes\nu}{Q_\lambda}
 &=\J_N(\lambda)-H(X)=B_N+C_N(\lambda),
 \label{eq:augmented_identity}\\
 \KL{P}{\pi_\lambda}
 &=B_N+C_N(\lambda)-R_N(\lambda),
 \label{eq:marginal_gap}
\end{align}
where $B_N=\sum_k\delta_k h(a_{k-1})-\int_0^1h\ge0$ is the expected conditional total correlation among tokens revealed in the same round, $C_N(\lambda)=\sum_k\delta_k[c_\lambda-h](a_{k-1})\ge0$ is the accumulated posterior error, and $R_N(\lambda)=\E_{X\sim P}\KL{\nu}{Q_\lambda(S\mid X)}\ge0$ is the credit for mixing paths.
\end{theorem}
\label{eq:nelbo}\label{eq:profile_accounting}
Conditioned on its reveal round $k$, a coordinate sees each other coordinate independently with probability $a_{k-1}$, so the path-augmented log-loss is exactly the left Riemann sum $\J_N$. Applying the chain rule for KL in the other order, conditioning on the output first, peels off the path credit $R_N$ (\Cref{app:accounting_proof}).
Each term has a distinct origin. $C_N$ is posterior error: it vanishes for the Bayes denoiser at $\lambda=1$, and it is what training and held-out loss measure. $B_N$ is the price of parallelism. It depends only on the data and the grid, so a perfect model pays it in full. Because $h$ is nonincreasing, it is the overshoot of a left Riemann sum, bounded by $\max_k\delta_k\sum_iI(X_i;X_{-i})$ and decaying like $1/N$ (\Cref{app:dependence_profile}). Generation, however, is judged on the mixture, and the mixing credit $R_N$ depends on $\lambda$. Minimizing posterior error therefore need not minimize output error. For two tokens the direction is explicit: an exact model whose tokens share a round with probability $c=\sum_k\delta_k^2$ has output entropy $H(\pi_1)\ge H(P)+c\,I(X_1;X_2)$, so parallel sampling makes a perfect model too random (\Cref{app:pair_entropy}). Many steps do not remove the effect: with $M=L=1024$, 63\% of tokens still share their round with another token (\Cref{app:occupancy}).

\subsection{Temperature one is almost never right for parallel sampling}
\label{sec:generic}
\begin{theorem}[The Bayes temperature is generically suboptimal]
\label{thm:generic}
Let $q$ be the Bayes denoiser, $q^i(\cdot\mid x_A)=P(X_i=\cdot\mid X_A=x_A)$, with $L\ge2$ and $|\mathcal V|\ge2$.
\textbf{(a)} If every round reveals at most one coordinate, in any order independent of $X$, then $\pi_1=P$, so $\lambda=1$ is optimal.
\textbf{(b)} For the parallel sampler with any grid of $N\ge2$ rounds, the full-support targets $P$ for which $\lambda=1$ is a stationary point of $\lambda\mapsto\KL{P}{\pi_\lambda}$ form a closed set of Lebesgue measure zero. For every other target, some $\lambda\neq1$ gives the Bayes denoiser a strictly better output law.
\end{theorem}
Part (a) is the chain rule. For (b), the derivative at $\lambda=1$ is analytic in $P$, so its zero set is null~\citep{mityagin2015zero} unless it vanishes identically, which a $|\mathcal V|$-ary pair family rules out (\Cref{app:generic_proof}).

The theorem separates parallel sampling from autoregressive decoding, where temperature one already gives the exact output law for an exact model. The theorem does not fix the direction, but the derivative has an explicit covariance form (\Cref{app:generic_proof}), and exhaustive enumeration finds sharpening to be the improving direction for 279 of 280 random small targets (\Cref{fig:theory}c). The pair bound of \Cref{sec:entropy_profile} suggests why: same-round draws add randomness that the target does not have, and sharpening removes it. Temperature is thus not a patch for a miscalibrated model. It is a free parameter of the parallel sampler that the Bayes posterior leaves unset.

\subsection{Exact repair: a worse predictor can generate better}
\label{sec:collision}
Two uniform bits that agree with probability $s$ make the mechanism exact:
\begin{equation}
 P_s(00)=P_s(11)=s/2,\qquad P_s(01)=P_s(10)=(1-s)/2,
 \qquad \tfrac12<s<1.
 \label{eq:binary_target}
\end{equation}
The tokens share a round with probability $c=\sum_k\delta_k^2$. That branch draws two independent uniform bits, which agree only half the time. Otherwise one token is revealed first and the other is drawn from its conditional, which the tempered Bayes denoiser makes agree with probability $u_\lambda=\operatorname{sigmoid}(\lambda\logit s)$.

\begin{proposition}[Exact repair and inverted rankings]
\label{prop:repair}\label{thm:compensation}
Let $0<c<1$. The sampled law is $P_{t_\lambda}$ with $t_\lambda=c/2+(1-c)u_\lambda$, and
\begin{equation}
 \KL{P_s}{\pi_\lambda}=\KL{\Bern(s)}{\Bern(t_\lambda)},\qquad
 C_N(\lambda)=(1-c)\KL{\Bern(s)}{\Bern(u_\lambda)}.
 \label{eq:binary_objectives}
\end{equation}
\textbf{(i) Exact repair.} If $c<2(1-s)$, let $u_\star=(s-c/2)/(1-c)$. Then $\lambda_\star=\logit(u_\star)/\logit(s)>1$ makes the sampler exact, $\pi_{\lambda_\star}=P_s$, although $C_N(\lambda_\star)>0=C_N(1)$. If $c\ge2(1-s)$, output error decreases with every increase of $\lambda$.
\textbf{(ii) Inverted ranking.} In the finite-repair regime, any denoiser with uniform marginals and conditional agreement $u\in(s,u_\star]$, sampled at $\lambda=1$, has strictly larger posterior loss and strictly smaller output error than the Bayes denoiser.
\textbf{(iii) Vanishing with steps.} As $c\to0$, $\lambda_\star=1+\frac{s-1/2}{s(1-s)\logit s}\,c+O(c^2)$, where $c=1/N$ for $N$ equal rounds.
\end{proposition}
\label{eq:exact_repair}\label{eq:main_collision_expansion}

With two equal rounds and $s=.7$, the independent branch agrees with probability $.5$ and the sequential branch with $.7$, so the mixture agrees with probability $.6$: too random. Sharpening to $\lambda_\star\approx2.59$ raises sequential agreement to $.9$, restoring the target $.7$ exactly (\Cref{fig:theory}a,b). In the accounting identity, $C_N$ rises but $R_N$ rises by more: improving the mixture requires worsening its components. Nothing here is special to temperature. Any continuous sharpening family that moves sequential agreement from $s$ toward $1$ passes through $u_\star$ and repairs the sampler; truncation-style sharpening, including fp32 Gumbel noise, pulls the same lever. Part (ii) is the inversion of \Cref{sec:less} in miniature: the less accurate predictor is the better few-step generator. Part (iii) explains why sharpening matters most at few steps and fades as rounds become sequential. The strict benefit survives small perturbations of the target and grid (\Cref{app:compensation_proof}).

\subsection{In trained models, sharpening hurts prediction and helps generation}
\label{sec:certificate}
If sharpening compensates for parallel sampling rather than for an underconfident model, it should improve samples while \emph{worsening} the schedule-matched posterior loss $\J_N$ of \Cref{thm:accounting}. We estimate $\J_N$ without bias on 512 held-out OpenWebText sequences, masking one target token and drawing its context from the reveal grid (\Cref{app:temperature_proof}). Moving from $\tau=1$ to $.9$ raises $\J_{M+1}/L$ ($M$ updates plus a final cleanup round) by $.036$--$.068$ nats per token for three MDLM-Small checkpoints at five budgets, with all 15 paired intervals above zero, and the posterior-optimal temperature on our grid is $1$ or $1.05$ (\Cref{tab:certificate_measurement}). Over the same cells, gPPL falls $2.6$--$3.8\times$ (\Cref{fig:overview}c). The checkpoint inversion of \Cref{sec:less} carries the same signature: the 915k-step checkpoint has lower $\J_N$ at every budget and a lower posterior-loss profile at every nontrivial clean-probability level (\Cref{app:certificate_measurement}), yet is the worse few-step generator at the standard sampler. Held-out loss sees only $B_N+C_N$; the samples also see $R_N$, so held-out loss cannot certify any sampler, ours included (\Cref{app:samplers}). The theory shows that the mechanism exists and is generic, not how the language-model gain splits among parallelization, evaluator preferences, and precision.

%% file: sections/conclusion.tex
\section{Discussion and Conclusion}
\label{sec:discussion}
We uncover how a masked diffusion language model from a few years ago, sampled slightly differently, rivals models distilled to replace it and gains quality and diversity together under group-level judging. Our findings have significant implications for the broader DLM landscape.

\paragraph{A new method should beat the baseline's best sampler.}
A gain over a default configuration mixes a new method's contribution with what the baseline sampler leaves on the table. Tune temperature and categorical precision for both under the same protocol, and report the search budget and the sampler behind each result. Sharpening can also be implicit, through categorical precision or a selection rule, and configured steps should be reported alongside network calls and measured costs.

\paragraph{Better predictions need not produce better generators.}
Training reduces posterior error but cannot restore the dependence lost when same-round tokens are drawn independently. Sharpening compensates for this loss, better schedules and joint draws reduce its source, and revision revisits committed tokens (\Cref{thm:accounting,app:samplers}).

\paragraph{Generation quality requires comparisons across outputs.}
\textsc{GEQ} and \textsc{GED} report quality and across-output diversity side by side under one fixed protocol (\Cref{prop:collapse} shows why both are needed). On outputs studied, they confirm MDLM's sharpening gain, find no judge-robust gain from sharpening DiDi-Instruct, and show that DiDi-Instruct's perplexity lead is not a judged quality lead.

\paragraph{Scope and limitations.}
Our central audit concerns unconditional generation with small masked diffusion models, albeit very popular ones. The cross-model comparison adds DUO, SEDD, and DiDi-Instruct, but tunes the sampler only for MDLM and DiDi-Instruct, so it does not show how far sharpening would move the other families. The tested larger 8B models already use greedy decoding, and positive temperatures do not help (\Cref{app:large_models}). The theory assumes exact, non-adaptive reveal sampling, and adaptive and revising samplers need further analysis. Comprehensive human validation of \textsc{GroupEval}'s quality scores and mode assignments would add further value.

The sampler is part of the model. A checkpoint is only as good as the sampler it is deployed with, and its training loss cannot say which sampler that should be. Once the sampler is tuned, an old baseline stands with the methods built to replace it. Once outputs are compared with each other, a large perplexity lead can turn out to carry no quality lead. Both corrections come from measuring the deployed generator rather than its predictions or single samples. To identify progress in few-step generation, tune the sampler and evaluate the quality and range of what it produces.

%% file: ai_use_statement.tex
\section*{AI use statement}
Generative AI tools assisted with literature retrieval, refinement of the theoretical framework, and creating presentation figures. The authors remain responsible for the final text and artifacts.

%% file: sections/app_exp.tex
\section{Additional Experiments and Implementation Details}
\label{app:experiments}
% This appendix provides detailed experimental setup, full sampling grids, and supporting comparisons, followed by the large-model stress test.

\subsection{Experimental setup}
\label{app:setup}

This subsection consolidates dataset, model, sampler, and evaluation details for the reported experiments. Throughout, the deployed object is the sampler--model pair \(\pi_0^{\theta,M,\tau,\kappa}\), where \(\theta\) is the denoiser checkpoint, \(M\) is the configured number of reverse-update invocations, \(\tau\) is the explicit sampling temperature, and \(\kappa\in\{\mathrm{fp32},\mathrm{fp64}\}\) is the categorical-kernel precision. Implementations also apply their configured terminal noise removal. Because ancestral caching can reuse logits on empty updates, actual model-call counts are path-dependent and were not instrumented in the quality grids. Note that \(M\) is not NFE, and cross-model rows are not compute-matched (the exact update/NFE relationship is derived in \Cref{app:implementation}). We call \((\tau{=}1,\mathrm{fp64})\) the \emph{standard} sampler and \((\tau{=}0.9,\mathrm{fp32})\) the \emph{sharpened} sampler; when applied to the public MDLM-Small checkpoint, the latter is \sharpm{}. Released aggregates label the two settings \texttt{reference} and \texttt{selected}. The sharpened sampler is an empirical, guard-constrained operating point, instead of a universal recommendation or a posterior-calibration optimum.

\paragraph{Data.}
The small-model unconditional-generation experiments use OpenWebText~\citep{gokaslan2019openwebtext} with the GPT-2 BPE tokenizer~\citep{radford2019gpt2}. Unless stated otherwise, generation is unconditional at sequence length \(L{=}1024\) tokens. The 128-sample MDLM grids use GPT-2 as the external evaluator. A controlled exact-output sensitivity study additionally applies GPT-2-Large and Qwen3-8B-Base evaluators under common span semantics (\Cref{tab:groupeval-exact40-common}).

\paragraph{Models.}
We train two MDLM variants from the official codebase of~\citet{sahoo2024simple}, denoted \textsc{MDLM-Tiny} and \textsc{MDLM-Small}, on OpenWebText with the standard SUBS parameterization, absorbing forward process, and continuous-time NELBO objective (\Cref{eq:nelbo}); training hyperparameters follow~\citet{sahoo2024simple} unless noted. \textsc{MDLM-Small} matches the architecture and parameter count of the publicly released MDLM-Small checkpoint of~\citet{sahoo2024simple}; \textsc{MDLM-Tiny} is a smaller variant retaining the same architecture family. The reported MDLM-Small early and late checkpoint identifiers are \texttt{0-20788.ckpt} and \texttt{8-914672.ckpt}, corresponding to near-completion of first and ninth epoch. The \emph{public} checkpoint is the much longer trained MDLM-Small release of~\citet{sahoo2024simple}, used as-is.

\paragraph{Cross-model checkpoints.}
For \Cref{tab:final_model_compare}, one consolidated artifact contains 40-sample GPT-2-Large evaluations of released checkpoints for MDLM~\citep{sahoo2024simple}, DUO and DUO-DCD~\citep{sahoo2025duo}, SEDD~\citep{lou2024sedd}, and DiDi-Instruct~\citep{zheng2025didi}. DiDi-Instruct uses the EMA student with its reward-guided ancestral sampler in the main-text comparison, while an unguided ancestral grid is also reported in \Cref{tab:didi_full_grid_ancestral}.

\paragraph{Sampler.}
We use the MDLM ancestral sampler with a linear \(\alpha\)-schedule and uniform time grid. Temperature is applied as post-hoc rescaling of the per-position \(x_0\) marginals (\Cref{eq:temperature_calibration}).
\(\kappa\in\{\mathrm{fp32},\mathrm{fp64}\}\) selects the floating-point precision of the Gumbel-max categorical draw. We sweep
\(\tau\in\{0.80,0.85,0.90,0.95,1.00,1.05,1.10\}\) and \(\kappa\in\{\mathrm{fp32},\mathrm{fp64}\}\), at configured update budgets \(M\in\{1,2,4,8,16,32,64,128,256,512,1024,2048\}\). 
% A subset is reported per table for compactness; full grids for our trained MDLM checkpoints are in \Cref{tab:mdlm_tiny_late_full_grid,tab:mdlm_small_late_full_grid,tab:mdlm_small_early_full_grid}, and for DiDi-Instruct in \Cref{tab:didi_full_grid_ancestral,tab:didi_full_grid_guided}.

% \paragraph{Sample budget per cell.}
% Each headline MDLM grid cell uses \(128\) unconditional samples; \textsc{MDLM-Tiny} full-budget cells use \(256\). The DiDi-Instruct grids and cross-model frontier (\Cref{tab:final_model_compare}) use \(40\) unconditional samples per point. We therefore present the cross-model result as a qualitative stress test rather than a definitive benchmark and report sample entropy alongside gPPL.

\paragraph{Evaluation metrics.}
Likelihood-based evaluation uses generative perplexity (gPPL, lower is better), with the causal-LM evaluator identified for each experiment. The compact aggregates pair it with within-sequence empirical unigram entropy \(H\) in nats. We use a declared conservative operating guard \(H\ge5\): cells below it are marked with \(\dagger\) and treated as aggressive Pareto points. This statistic exposes obvious within-sequence concentration but does not measure across-sample support coverage and is not a distributional-correctness certificate.

\paragraph{Hardware.}
Local generation and sampling-cost measurements use a single NVIDIA H100 (80 GB) GPU. GroupEval judges are documented separately. The matching cost comparison in \Cref{fig:cost_main} uses batch size \(8\), one warm-up batch, and three measured batches (24 sequences) per point under one software environment.

\paragraph{Reproducibility.}
Sampling temperature and categorical-kernel precision are the only knobs varied between standard and sharpened rows. Model weights, time grids, and the reveal rule are not changed at all. We do not retrain a model, fine-tune a sampler, or distill a student whatsoever.

% \clearpage
\input{sections/baseline_grids}
% \clearpage

\subsection{MDLM-Tiny: the same effect at smaller scale}
\label{app:tiny}

\begin{figure}[ht]
    \centering
    \includegraphics[width=0.55\linewidth]{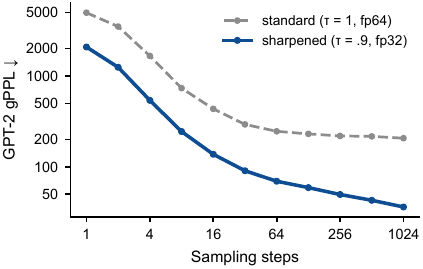}
    \caption{\textbf{MDLM-Tiny shows the same sampler gap at smaller scale.} Late MDLM-Tiny checkpoint; standard \((\tau{=}1,\mathrm{fp64})\), dashed; sharpened \((\tau{=}.9,\mathrm{fp32})\), solid. The gap mirrors the MDLM-Small headline (\Cref{tab:headline_main}).}
    \label{fig:tiny_calibration_appendix}
\end{figure}

\Cref{fig:tiny_calibration_appendix} reports the corresponding operating-point comparison for MDLM-Tiny. The same standard-versus-sharpened contrast holds: at every \(M\), the sharpened cell improves gPPL by a substantial multiple, with \(H\ge5\) over most of the measured budget range. The full MDLM-Tiny grid is in \Cref{tab:mdlm_tiny_late_full_grid}.

\subsection{Full sampling-configuration grids}
\label{app:full_grids}

In each grid below, columns are the seven temperatures \(\tau\in\{0.80,\dots,1.10\}\) and rows are configured reverse updates \(M\). Both sub-blocks are shown (fp64 above, fp32 below). To aid scanning, the standard \((\tau{=}1,\mathrm{fp64})\) and sharpened \((\tau{=}0.9,\mathrm{fp32})\) cells are emphasized in \textbf{bold}. Cells below the \(H<5\) operating guard are marked with \(\dagger\). Entries are gPPL with within-sequence entropy in brackets.

\begin{table}[!ht]
    \centering
    \scriptsize
    \caption{\textsc{MDLM-Tiny} late checkpoint, full operating-point grid. The standard \((\tau{=}1,\mathrm{fp64})\) and sharpened \((\tau{=}0.9,\mathrm{fp32})\) cells are in \textbf{bold}; \(\dagger\) marks \(H<5\).}
    \label{tab:mdlm_tiny_late_full_grid}
\setlength{\tabcolsep}{2.5pt}
\resizebox{\linewidth}{!}{%
\begin{tabular}{rccccccc}
\toprule
$M$ & $\tau{=}0.8$ & $\tau{=}0.85$ & $\tau{=}0.9$ & $\tau{=}0.95$ & $\tau{=}1.0$ & $\tau{=}1.05$ & $\tau{=}1.10$ \\
\midrule
\multicolumn{8}{l}{\textbf{fp64 categorical sampling.}} \\
$1$ & $786.0$\textsuperscript{$\dagger$} {\scriptsize[4.55]} & $1276$\textsuperscript{$\dagger$} {\scriptsize[4.92]} & $2050$ {\scriptsize[5.27]} & $3246$ {\scriptsize[5.58]} & $\mathbf{4961}$ {\scriptsize[5.84]} & $7231$ {\scriptsize[6.05]} & $10037$ {\scriptsize[6.23]} \\
$2$ & $428.7$\textsuperscript{$\dagger$} {\scriptsize[4.57]} & $732.8$\textsuperscript{$\dagger$} {\scriptsize[4.95]} & $1263$ {\scriptsize[5.32]} & $2132$ {\scriptsize[5.64]} & $\mathbf{3496}$ {\scriptsize[5.91]} & $5464$ {\scriptsize[6.13]} & $8034$ {\scriptsize[6.31]} \\
$4$ & $199.9$\textsuperscript{$\dagger$} {\scriptsize[4.63]} & $323.9$\textsuperscript{$\dagger$} {\scriptsize[4.99]} & $561.9$ {\scriptsize[5.33]} & $975.9$ {\scriptsize[5.62]} & $\mathbf{1661}$ {\scriptsize[5.87]} & $2807$ {\scriptsize[6.10]} & $4607$ {\scriptsize[6.28]} \\
$8$ & $99.0$\textsuperscript{$\dagger$} {\scriptsize[4.76]} & $154.0$ {\scriptsize[5.07]} & $262.9$ {\scriptsize[5.36]} & $440.5$ {\scriptsize[5.61]} & $\mathbf{734.1}$ {\scriptsize[5.81]} & $1283$ {\scriptsize[6.01]} & $2261$ {\scriptsize[6.19]} \\
$16$ & $59.9$\textsuperscript{$\dagger$} {\scriptsize[4.88]} & $93.1$ {\scriptsize[5.14]} & $151.3$ {\scriptsize[5.37]} & $253.7$ {\scriptsize[5.59]} & $\mathbf{435.1}$ {\scriptsize[5.78]} & $748.9$ {\scriptsize[5.95]} & $1371$ {\scriptsize[6.13]} \\
$32$ & $44.1$\textsuperscript{$\dagger$} {\scriptsize[4.91]} & $67.0$ {\scriptsize[5.15]} & $104.9$ {\scriptsize[5.36]} & $175.9$ {\scriptsize[5.56]} & $\mathbf{293.3}$ {\scriptsize[5.73]} & $508.6$ {\scriptsize[5.90]} & $910.6$ {\scriptsize[6.07]} \\
$64$ & $35.9$\textsuperscript{$\dagger$} {\scriptsize[4.86]} & $55.2$ {\scriptsize[5.11]} & $87.5$ {\scriptsize[5.33]} & $144.9$ {\scriptsize[5.53]} & $\mathbf{246.4}$ {\scriptsize[5.71]} & $441.5$ {\scriptsize[5.87]} & $805.4$ {\scriptsize[6.04]} \\
$128$ & $32.8$\textsuperscript{$\dagger$} {\scriptsize[4.83]} & $50.4$ {\scriptsize[5.09]} & $77.4$ {\scriptsize[5.28]} & $129.3$ {\scriptsize[5.50]} & $\mathbf{230.7}$ {\scriptsize[5.69]} & $418.8$ {\scriptsize[5.87]} & $747.6$ {\scriptsize[6.04]} \\
$256$ & $32.1$\textsuperscript{$\dagger$} {\scriptsize[4.82]} & $49.6$ {\scriptsize[5.08]} & $77.8$ {\scriptsize[5.30]} & $127.9$ {\scriptsize[5.51]} & $\mathbf{219.0}$ {\scriptsize[5.68]} & $377.5$ {\scriptsize[5.85]} & $691.4$ {\scriptsize[6.03]} \\
$512$ & $29.9$\textsuperscript{$\dagger$} {\scriptsize[4.80]} & $46.7$ {\scriptsize[5.06]} & $74.6$ {\scriptsize[5.29]} & $124.0$ {\scriptsize[5.49]} & $\mathbf{216.6}$ {\scriptsize[5.68]} & $384.0$ {\scriptsize[5.86]} & $725.4$ {\scriptsize[6.03]} \\
$1024$ & $30.2$\textsuperscript{$\dagger$} {\scriptsize[4.81]} & $44.8$ {\scriptsize[5.04]} & $71.5$ {\scriptsize[5.26]} & $118.0$ {\scriptsize[5.47]} & $\mathbf{207.0}$ {\scriptsize[5.66]} & $376.4$ {\scriptsize[5.85]} & $693.3$ {\scriptsize[6.02]} \\
\midrule
\multicolumn{8}{l}{\textbf{fp32 categorical sampling.}} \\
$1$ & $789.4$\textsuperscript{$\dagger$} {\scriptsize[4.55]} & $1288$\textsuperscript{$\dagger$} {\scriptsize[4.92]} & $\mathbf{2083}$ {\scriptsize[5.27]} & $3249$ {\scriptsize[5.58]} & $4922$ {\scriptsize[5.83]} & $7249$ {\scriptsize[6.05]} & $10089$ {\scriptsize[6.22]} \\
$2$ & $422.2$\textsuperscript{$\dagger$} {\scriptsize[4.55]} & $723.3$\textsuperscript{$\dagger$} {\scriptsize[4.94]} & $\mathbf{1246}$ {\scriptsize[5.31]} & $2120$ {\scriptsize[5.63]} & $3451$ {\scriptsize[5.90]} & $5387$ {\scriptsize[6.13]} & $7818$ {\scriptsize[6.30]} \\
$4$ & $193.2$\textsuperscript{$\dagger$} {\scriptsize[4.62]} & $323.3$\textsuperscript{$\dagger$} {\scriptsize[4.98]} & $\mathbf{540.0}$ {\scriptsize[5.32]} & $948.3$ {\scriptsize[5.62]} & $1594$ {\scriptsize[5.87]} & $2747$ {\scriptsize[6.09]} & $4504$ {\scriptsize[6.28]} \\
$8$ & $95.6$\textsuperscript{$\dagger$} {\scriptsize[4.74]} & $150.3$ {\scriptsize[5.07]} & $\mathbf{245.0}$ {\scriptsize[5.35]} & $416.4$ {\scriptsize[5.59]} & $691.5$ {\scriptsize[5.81]} & $1213$ {\scriptsize[6.00]} & $2142$ {\scriptsize[6.19]} \\
$16$ & $56.3$\textsuperscript{$\dagger$} {\scriptsize[4.84]} & $86.4$ {\scriptsize[5.12]} & $\mathbf{137.7}$ {\scriptsize[5.35]} & $221.6$ {\scriptsize[5.56]} & $368.7$ {\scriptsize[5.74]} & $654.6$ {\scriptsize[5.92]} & $1171$ {\scriptsize[6.09]} \\
$32$ & $38.2$\textsuperscript{$\dagger$} {\scriptsize[4.82]} & $56.7$ {\scriptsize[5.07]} & $\mathbf{90.7}$ {\scriptsize[5.30]} & $148.7$ {\scriptsize[5.50]} & $249.6$ {\scriptsize[5.68]} & $433.9$ {\scriptsize[5.85]} & $781.8$ {\scriptsize[6.03]} \\
$64$ & $31.5$\textsuperscript{$\dagger$} {\scriptsize[4.80]} & $45.0$ {\scriptsize[5.02]} & $\mathbf{69.6}$ {\scriptsize[5.24]} & $113.7$ {\scriptsize[5.44]} & $193.9$ {\scriptsize[5.63]} & $329.8$ {\scriptsize[5.80]} & $567.2$ {\scriptsize[5.97]} \\
$128$ & $27.9$\textsuperscript{$\dagger$} {\scriptsize[4.74]} & $40.0$\textsuperscript{$\dagger$} {\scriptsize[4.99]} & $\mathbf{59.2}$ {\scriptsize[5.19]} & $90.3$ {\scriptsize[5.39]} & $148.3$ {\scriptsize[5.57]} & $239.7$ {\scriptsize[5.74]} & $414.8$ {\scriptsize[5.90]} \\
$256$ & $24.6$\textsuperscript{$\dagger$} {\scriptsize[4.71]} & $34.5$\textsuperscript{$\dagger$} {\scriptsize[4.93]} & $\mathbf{49.7}$ {\scriptsize[5.12]} & $77.5$ {\scriptsize[5.32]} & $118.4$ {\scriptsize[5.49]} & $186.8$ {\scriptsize[5.66]} & $311.3$ {\scriptsize[5.83]} \\
$512$ & $22.1$\textsuperscript{$\dagger$} {\scriptsize[4.63]} & $30.3$\textsuperscript{$\dagger$} {\scriptsize[4.85]} & $\mathbf{42.9}$ {\scriptsize[5.05]} & $61.7$ {\scriptsize[5.22]} & $94.7$ {\scriptsize[5.40]} & $147.2$ {\scriptsize[5.57]} & $237.5$ {\scriptsize[5.74]} \\
$1024$ & $20.5$\textsuperscript{$\dagger$} {\scriptsize[4.58]} & $27.2$\textsuperscript{$\dagger$} {\scriptsize[4.78]} & $\mathbf{36.3}$\textsuperscript{$\dagger$} {\scriptsize[4.94]} & $50.7$ {\scriptsize[5.11]} & $72.9$ {\scriptsize[5.29]} & $107.1$ {\scriptsize[5.45]} & $165.1$ {\scriptsize[5.62]} \\
\bottomrule
\end{tabular}}
\end{table}

\begin{table}[!ht]
    \centering
    \scriptsize
    \caption{\textsc{MDLM-Small} late checkpoint, full operating-point grid. The standard \((\tau{=}1,\mathrm{fp64})\) and sharpened \((\tau{=}0.9,\mathrm{fp32})\) cells are in \textbf{bold}; \(\dagger\) marks \(H<5\).}
    \label{tab:mdlm_small_late_full_grid}
\setlength{\tabcolsep}{2.5pt}
\resizebox{\linewidth}{!}{%
\begin{tabular}{rccccccc}
\toprule
$M$ & $\tau{=}0.8$ & $\tau{=}0.85$ & $\tau{=}0.9$ & $\tau{=}0.95$ & $\tau{=}1.0$ & $\tau{=}1.05$ & $\tau{=}1.10$ \\
\midrule
\multicolumn{8}{l}{\textbf{fp64 categorical sampling.}} \\
$1$ & $759.3$\textsuperscript{$\dagger$} {\scriptsize[4.52]} & $1244$\textsuperscript{$\dagger$} {\scriptsize[4.90]} & $1992$ {\scriptsize[5.25]} & $3135$ {\scriptsize[5.55]} & $\mathbf{4780}$ {\scriptsize[5.82]} & $6957$ {\scriptsize[6.04]} & $9722$ {\scriptsize[6.21]} \\
$2$ & $426.1$\textsuperscript{$\dagger$} {\scriptsize[4.51]} & $734.8$\textsuperscript{$\dagger$} {\scriptsize[4.92]} & $1280$ {\scriptsize[5.31]} & $2181$ {\scriptsize[5.64]} & $\mathbf{3634}$ {\scriptsize[5.93]} & $5780$ {\scriptsize[6.16]} & $8385$ {\scriptsize[6.34]} \\
$4$ & $179.8$\textsuperscript{$\dagger$} {\scriptsize[4.54]} & $333.0$\textsuperscript{$\dagger$} {\scriptsize[4.96]} & $600.2$ {\scriptsize[5.34]} & $1075$ {\scriptsize[5.66]} & $\mathbf{1931}$ {\scriptsize[5.95]} & $3344$ {\scriptsize[6.18]} & $5612$ {\scriptsize[6.38]} \\
$8$ & $86.3$\textsuperscript{$\dagger$} {\scriptsize[4.67]} & $141.0$ {\scriptsize[5.03]} & $236.4$ {\scriptsize[5.35]} & $421.5$ {\scriptsize[5.62]} & $\mathbf{795.6}$ {\scriptsize[5.88]} & $1561$ {\scriptsize[6.10]} & $2938$ {\scriptsize[6.30]} \\
$16$ & $45.3$\textsuperscript{$\dagger$} {\scriptsize[4.82]} & $66.8$ {\scriptsize[5.12]} & $107.6$ {\scriptsize[5.38]} & $186.6$ {\scriptsize[5.59]} & $\mathbf{333.9}$ {\scriptsize[5.78]} & $615.9$ {\scriptsize[5.97]} & $1262$ {\scriptsize[6.17]} \\
$32$ & $29.2$\textsuperscript{$\dagger$} {\scriptsize[4.90]} & $41.3$ {\scriptsize[5.14]} & $64.7$ {\scriptsize[5.35]} & $106.1$ {\scriptsize[5.54]} & $\mathbf{194.5}$ {\scriptsize[5.72]} & $355.8$ {\scriptsize[5.91]} & $761.4$ {\scriptsize[6.10]} \\
$64$ & $23.1$\textsuperscript{$\dagger$} {\scriptsize[4.88]} & $34.4$ {\scriptsize[5.13]} & $52.3$ {\scriptsize[5.34]} & $84.0$ {\scriptsize[5.51]} & $\mathbf{144.0}$ {\scriptsize[5.69]} & $266.8$ {\scriptsize[5.86]} & $573.7$ {\scriptsize[6.04]} \\
$128$ & $20.3$\textsuperscript{$\dagger$} {\scriptsize[4.88]} & $29.0$ {\scriptsize[5.11]} & $44.3$ {\scriptsize[5.31]} & $71.9$ {\scriptsize[5.50]} & $\mathbf{121.4}$ {\scriptsize[5.67]} & $225.8$ {\scriptsize[5.83]} & $461.7$ {\scriptsize[6.01]} \\
$256$ & $18.8$\textsuperscript{$\dagger$} {\scriptsize[4.82]} & $27.9$ {\scriptsize[5.08]} & $42.4$ {\scriptsize[5.30]} & $70.1$ {\scriptsize[5.48]} & $\mathbf{122.1}$ {\scriptsize[5.66]} & $237.8$ {\scriptsize[5.84]} & $496.5$ {\scriptsize[6.03]} \\
$512$ & $17.9$\textsuperscript{$\dagger$} {\scriptsize[4.79]} & $25.5$ {\scriptsize[5.05]} & $38.0$ {\scriptsize[5.26]} & $63.7$ {\scriptsize[5.46]} & $\mathbf{111.9}$ {\scriptsize[5.64]} & $207.7$ {\scriptsize[5.82]} & $425.9$ {\scriptsize[6.00]} \\
$1024$ & $18.1$\textsuperscript{$\dagger$} {\scriptsize[4.84]} & $25.5$ {\scriptsize[5.07]} & $38.6$ {\scriptsize[5.26]} & $62.9$ {\scriptsize[5.46]} & $\mathbf{108.2}$ {\scriptsize[5.63]} & $189.0$ {\scriptsize[5.79]} & $384.8$ {\scriptsize[5.97]} \\
\midrule
\multicolumn{8}{l}{\textbf{fp32 categorical sampling.}} \\
$1$ & $785.0$\textsuperscript{$\dagger$} {\scriptsize[4.52]} & $1249$\textsuperscript{$\dagger$} {\scriptsize[4.90]} & $\mathbf{2045}$ {\scriptsize[5.25]} & $3197$ {\scriptsize[5.55]} & $4733$ {\scriptsize[5.82]} & $6887$ {\scriptsize[6.04]} & $9636$ {\scriptsize[6.21]} \\
$2$ & $396.9$\textsuperscript{$\dagger$} {\scriptsize[4.48]} & $704.9$\textsuperscript{$\dagger$} {\scriptsize[4.90]} & $\mathbf{1219}$ {\scriptsize[5.29]} & $2104$ {\scriptsize[5.64]} & $3513$ {\scriptsize[5.92]} & $5488$ {\scriptsize[6.16]} & $8080$ {\scriptsize[6.34]} \\
$4$ & $174.3$\textsuperscript{$\dagger$} {\scriptsize[4.51]} & $312.9$\textsuperscript{$\dagger$} {\scriptsize[4.93]} & $\mathbf{589.6}$ {\scriptsize[5.32]} & $1040$ {\scriptsize[5.65]} & $1870$ {\scriptsize[5.94]} & $3351$ {\scriptsize[6.19]} & $5456$ {\scriptsize[6.38]} \\
$8$ & $81.2$\textsuperscript{$\dagger$} {\scriptsize[4.63]} & $132.1$ {\scriptsize[5.00]} & $\mathbf{231.8}$ {\scriptsize[5.32]} & $400.8$ {\scriptsize[5.60]} & $725.3$ {\scriptsize[5.84]} & $1396$ {\scriptsize[6.07]} & $2633$ {\scriptsize[6.27]} \\
$16$ & $42.7$\textsuperscript{$\dagger$} {\scriptsize[4.77]} & $66.3$ {\scriptsize[5.10]} & $\mathbf{96.7}$ {\scriptsize[5.35]} & $169.1$ {\scriptsize[5.57]} & $305.0$ {\scriptsize[5.76]} & $537.1$ {\scriptsize[5.94]} & $1166$ {\scriptsize[6.15]} \\
$32$ & $28.0$\textsuperscript{$\dagger$} {\scriptsize[4.89]} & $40.7$ {\scriptsize[5.14]} & $\mathbf{60.9}$ {\scriptsize[5.34]} & $93.1$ {\scriptsize[5.51]} & $151.2$ {\scriptsize[5.68]} & $283.7$ {\scriptsize[5.85]} & $560.5$ {\scriptsize[6.03]} \\
$64$ & $19.6$\textsuperscript{$\dagger$} {\scriptsize[4.78]} & $28.4$ {\scriptsize[5.04]} & $\mathbf{43.7}$ {\scriptsize[5.28]} & $65.7$ {\scriptsize[5.45]} & $105.5$ {\scriptsize[5.62]} & $181.0$ {\scriptsize[5.77]} & $365.7$ {\scriptsize[5.94]} \\
$128$ & $16.8$\textsuperscript{$\dagger$} {\scriptsize[4.76]} & $24.0$ {\scriptsize[5.00]} & $\mathbf{33.8}$ {\scriptsize[5.20]} & $51.2$ {\scriptsize[5.38]} & $80.1$ {\scriptsize[5.55]} & $128.6$ {\scriptsize[5.70]} & $243.1$ {\scriptsize[5.87]} \\
$256$ & $15.4$\textsuperscript{$\dagger$} {\scriptsize[4.70]} & $21.1$\textsuperscript{$\dagger$} {\scriptsize[4.95]} & $\mathbf{29.3}$ {\scriptsize[5.14]} & $41.8$ {\scriptsize[5.30]} & $64.3$ {\scriptsize[5.47]} & $107.8$ {\scriptsize[5.63]} & $181.5$ {\scriptsize[5.78]} \\
$512$ & $13.7$\textsuperscript{$\dagger$} {\scriptsize[4.63]} & $18.0$\textsuperscript{$\dagger$} {\scriptsize[4.84]} & $\mathbf{24.7}$ {\scriptsize[5.05]} & $35.4$ {\scriptsize[5.24]} & $50.1$ {\scriptsize[5.38]} & $80.1$ {\scriptsize[5.54]} & $121.0$ {\scriptsize[5.69]} \\
$1024$ & $12.9$\textsuperscript{$\dagger$} {\scriptsize[4.61]} & $16.7$\textsuperscript{$\dagger$} {\scriptsize[4.82]} & $\mathbf{22.5}$ {\scriptsize[5.03]} & $31.0$ {\scriptsize[5.18]} & $41.9$ {\scriptsize[5.33]} & $61.5$ {\scriptsize[5.46]} & $96.0$ {\scriptsize[5.60]} \\
\bottomrule
\end{tabular}}
\end{table}

\begin{table}[!ht]
    \centering
    \scriptsize
    \caption{\textsc{MDLM-Small} early checkpoint, full operating-point grid. The standard \((\tau{=}1,\mathrm{fp64})\) and sharpened \((\tau{=}0.9,\mathrm{fp32})\) cells are in \textbf{bold}; \(\dagger\) marks \(H<5\).}
    \label{tab:mdlm_small_early_full_grid}
\setlength{\tabcolsep}{2.5pt}
\resizebox{\linewidth}{!}{%
\begin{tabular}{rccccccc}
\toprule
$M$ & $\tau{=}0.8$ & $\tau{=}0.85$ & $\tau{=}0.9$ & $\tau{=}0.95$ & $\tau{=}1.0$ & $\tau{=}1.05$ & $\tau{=}1.10$ \\
\midrule
\multicolumn{8}{l}{\textbf{fp64 categorical sampling.}} \\
$1$ & $782.9$\textsuperscript{$\dagger$} {\scriptsize[4.48]} & $1251$\textsuperscript{$\dagger$} {\scriptsize[4.87]} & $2007$ {\scriptsize[5.22]} & $3162$ {\scriptsize[5.54]} & $\mathbf{4941}$ {\scriptsize[5.80]} & $7116$ {\scriptsize[6.03]} & $9854$ {\scriptsize[6.20]} \\
$2$ & $431.3$\textsuperscript{$\dagger$} {\scriptsize[4.55]} & $703.7$\textsuperscript{$\dagger$} {\scriptsize[4.94]} & $1212$ {\scriptsize[5.31]} & $2022$ {\scriptsize[5.62]} & $\mathbf{3278}$ {\scriptsize[5.89]} & $5088$ {\scriptsize[6.11]} & $7647$ {\scriptsize[6.29]} \\
$4$ & $189.0$\textsuperscript{$\dagger$} {\scriptsize[4.72]} & $305.5$ {\scriptsize[5.07]} & $507.2$ {\scriptsize[5.38]} & $835.1$ {\scriptsize[5.63]} & $\mathbf{1376}$ {\scriptsize[5.86]} & $2313$ {\scriptsize[6.07]} & $3891$ {\scriptsize[6.25]} \\
$8$ & $96.1$\textsuperscript{$\dagger$} {\scriptsize[4.90]} & $144.2$ {\scriptsize[5.17]} & $235.2$ {\scriptsize[5.41]} & $365.8$ {\scriptsize[5.61]} & $\mathbf{605.9}$ {\scriptsize[5.80]} & $1024$ {\scriptsize[5.99]} & $1801$ {\scriptsize[6.15]} \\
$16$ & $55.3$\textsuperscript{$\dagger$} {\scriptsize[4.94]} & $85.3$ {\scriptsize[5.19]} & $131.1$ {\scriptsize[5.38]} & $211.5$ {\scriptsize[5.57]} & $\mathbf{343.1}$ {\scriptsize[5.73]} & $581.9$ {\scriptsize[5.91]} & $1052$ {\scriptsize[6.07]} \\
$32$ & $39.7$\textsuperscript{$\dagger$} {\scriptsize[4.93]} & $56.5$ {\scriptsize[5.14]} & $88.1$ {\scriptsize[5.34]} & $146.1$ {\scriptsize[5.53]} & $\mathbf{251.2}$ {\scriptsize[5.70]} & $390.4$ {\scriptsize[5.86]} & $732.5$ {\scriptsize[6.02]} \\
$64$ & $33.0$\textsuperscript{$\dagger$} {\scriptsize[4.91]} & $49.7$ {\scriptsize[5.13]} & $75.1$ {\scriptsize[5.32]} & $123.8$ {\scriptsize[5.51]} & $\mathbf{222.1}$ {\scriptsize[5.69]} & $377.4$ {\scriptsize[5.85]} & $685.9$ {\scriptsize[6.02]} \\
$128$ & $29.9$\textsuperscript{$\dagger$} {\scriptsize[4.90]} & $44.8$ {\scriptsize[5.10]} & $70.2$ {\scriptsize[5.30]} & $111.8$ {\scriptsize[5.48]} & $\mathbf{190.0}$ {\scriptsize[5.67]} & $325.0$ {\scriptsize[5.83]} & $573.6$ {\scriptsize[5.99]} \\
$256$ & $28.5$\textsuperscript{$\dagger$} {\scriptsize[4.89]} & $43.4$ {\scriptsize[5.10]} & $68.8$ {\scriptsize[5.31]} & $110.7$ {\scriptsize[5.47]} & $\mathbf{185.3}$ {\scriptsize[5.64]} & $328.7$ {\scriptsize[5.83]} & $614.7$ {\scriptsize[6.00]} \\
$512$ & $26.8$\textsuperscript{$\dagger$} {\scriptsize[4.83]} & $38.8$ {\scriptsize[5.05]} & $60.1$ {\scriptsize[5.25]} & $98.1$ {\scriptsize[5.44]} & $\mathbf{180.1}$ {\scriptsize[5.64]} & $315.1$ {\scriptsize[5.81]} & $614.4$ {\scriptsize[5.99]} \\
$1024$ & $26.4$\textsuperscript{$\dagger$} {\scriptsize[4.84]} & $40.0$ {\scriptsize[5.07]} & $57.5$ {\scriptsize[5.23]} & $96.0$ {\scriptsize[5.43]} & $\mathbf{166.1}$ {\scriptsize[5.62]} & $292.0$ {\scriptsize[5.79]} & $519.3$ {\scriptsize[5.96]} \\
\midrule
\multicolumn{8}{l}{\textbf{fp32 categorical sampling.}} \\
$1$ & $798.5$\textsuperscript{$\dagger$} {\scriptsize[4.48]} & $1281$\textsuperscript{$\dagger$} {\scriptsize[4.87]} & $\mathbf{2085}$ {\scriptsize[5.23]} & $3275$ {\scriptsize[5.54]} & $4864$ {\scriptsize[5.80]} & $7041$ {\scriptsize[6.03]} & $9793$ {\scriptsize[6.21]} \\
$2$ & $404.8$\textsuperscript{$\dagger$} {\scriptsize[4.52]} & $686.8$\textsuperscript{$\dagger$} {\scriptsize[4.93]} & $\mathbf{1151}$ {\scriptsize[5.29]} & $1904$ {\scriptsize[5.61]} & $3105$ {\scriptsize[5.88]} & $4811$ {\scriptsize[6.11]} & $7258$ {\scriptsize[6.28]} \\
$4$ & $178.3$\textsuperscript{$\dagger$} {\scriptsize[4.71]} & $297.5$ {\scriptsize[5.06]} & $\mathbf{498.2}$ {\scriptsize[5.37]} & $815.3$ {\scriptsize[5.63]} & $1385$ {\scriptsize[5.86]} & $2304$ {\scriptsize[6.06]} & $3731$ {\scriptsize[6.24]} \\
$8$ & $91.7$\textsuperscript{$\dagger$} {\scriptsize[4.88]} & $138.6$ {\scriptsize[5.15]} & $\mathbf{213.7}$ {\scriptsize[5.39]} & $344.6$ {\scriptsize[5.59]} & $551.5$ {\scriptsize[5.78]} & $923.9$ {\scriptsize[5.96]} & $1665$ {\scriptsize[6.14]} \\
$16$ & $52.5$\textsuperscript{$\dagger$} {\scriptsize[4.93]} & $76.2$ {\scriptsize[5.16]} & $\mathbf{120.5}$ {\scriptsize[5.37]} & $195.1$ {\scriptsize[5.55]} & $324.6$ {\scriptsize[5.72]} & $531.1$ {\scriptsize[5.90]} & $956.4$ {\scriptsize[6.05]} \\
$32$ & $36.8$\textsuperscript{$\dagger$} {\scriptsize[4.92]} & $53.6$ {\scriptsize[5.13]} & $\mathbf{81.7}$ {\scriptsize[5.32]} & $131.1$ {\scriptsize[5.49]} & $205.1$ {\scriptsize[5.65]} & $349.4$ {\scriptsize[5.82]} & $586.1$ {\scriptsize[5.97]} \\
$64$ & $27.9$\textsuperscript{$\dagger$} {\scriptsize[4.81]} & $42.2$ {\scriptsize[5.06]} & $\mathbf{61.7}$ {\scriptsize[5.26]} & $96.3$ {\scriptsize[5.43]} & $152.1$ {\scriptsize[5.60]} & $263.3$ {\scriptsize[5.76]} & $442.0$ {\scriptsize[5.91]} \\
$128$ & $24.5$\textsuperscript{$\dagger$} {\scriptsize[4.78]} & $34.8$\textsuperscript{$\dagger$} {\scriptsize[4.99]} & $\mathbf{50.4}$ {\scriptsize[5.17]} & $76.7$ {\scriptsize[5.35]} & $120.6$ {\scriptsize[5.53]} & $201.7$ {\scriptsize[5.69]} & $345.3$ {\scriptsize[5.86]} \\
$256$ & $22.7$\textsuperscript{$\dagger$} {\scriptsize[4.74]} & $30.6$\textsuperscript{$\dagger$} {\scriptsize[4.92]} & $\mathbf{42.8}$ {\scriptsize[5.11]} & $63.1$ {\scriptsize[5.28]} & $94.0$ {\scriptsize[5.45]} & $154.7$ {\scriptsize[5.61]} & $246.6$ {\scriptsize[5.76]} \\
$512$ & $19.3$\textsuperscript{$\dagger$} {\scriptsize[4.63]} & $26.7$\textsuperscript{$\dagger$} {\scriptsize[4.84]} & $\mathbf{36.2}$ {\scriptsize[5.02]} & $50.8$ {\scriptsize[5.16]} & $75.3$ {\scriptsize[5.35]} & $112.1$ {\scriptsize[5.51]} & $177.5$ {\scriptsize[5.67]} \\
$1024$ & $18.5$\textsuperscript{$\dagger$} {\scriptsize[4.59]} & $24.0$\textsuperscript{$\dagger$} {\scriptsize[4.79]} & $\mathbf{32.1}$\textsuperscript{$\dagger$} {\scriptsize[4.95]} & $44.6$ {\scriptsize[5.13]} & $64.4$ {\scriptsize[5.30]} & $93.2$ {\scriptsize[5.44]} & $134.4$ {\scriptsize[5.58]} \\
\bottomrule
\end{tabular}}
\end{table}

\clearpage
\begin{table}[!ht]
    \centering
    \scriptsize
    \caption{\textbf{DiDi-Instruct, unguided ancestral sampling.} GPT-2-Large gPPL [$H$], 40 samples per cell. Bold: standard $(1,\mathrm{fp64})$ and sharpened $(.9,\mathrm{fp32})$; $\dagger$: $H<5$.}
    \label{tab:didi_full_grid_ancestral}
\setlength{\tabcolsep}{2.5pt}
\resizebox{.97\linewidth}{!}{%
\begin{tabular}{rccccccc}
\toprule
$M$ & $\tau{=}0.8$ & $\tau{=}0.85$ & $\tau{=}0.9$ & $\tau{=}0.95$ & $\tau{=}1.0$ & $\tau{=}1.05$ & $\tau{=}1.10$ \\
\midrule
\multicolumn{8}{l}{\textbf{fp64 categorical sampling.}} \\
$1$ & $506.6$\textsuperscript{$\dagger$} {\scriptsize[4.44]} & $832.6$\textsuperscript{$\dagger$} {\scriptsize[4.79]} & $1393$ {\scriptsize[5.15]} & $2268$ {\scriptsize[5.46]} & $\mathbf{3371}$ {\scriptsize[5.72]} & $4958$ {\scriptsize[5.95]} & $7007$ {\scriptsize[6.13]} \\
$2$ & $288.3$\textsuperscript{$\dagger$} {\scriptsize[4.26]} & $483.8$\textsuperscript{$\dagger$} {\scriptsize[4.70]} & $866.1$ {\scriptsize[5.12]} & $1538$ {\scriptsize[5.50]} & $\mathbf{2487}$ {\scriptsize[5.79]} & $3958$ {\scriptsize[6.05]} & $6114$ {\scriptsize[6.25]} \\
$4$ & $110.4$\textsuperscript{$\dagger$} {\scriptsize[4.38]} & $160.8$\textsuperscript{$\dagger$} {\scriptsize[4.80]} & $241.1$ {\scriptsize[5.10]} & $367.1$ {\scriptsize[5.36]} & $\mathbf{532.5}$ {\scriptsize[5.56]} & $757.3$ {\scriptsize[5.72]} & $1168$ {\scriptsize[5.90]} \\
$8$ & $46.07$\textsuperscript{$\dagger$} {\scriptsize[4.72]} & $57.75$\textsuperscript{$\dagger$} {\scriptsize[4.94]} & $77.23$ {\scriptsize[5.15]} & $106.9$ {\scriptsize[5.31]} & $\mathbf{148.2}$ {\scriptsize[5.46]} & $192.7$ {\scriptsize[5.58]} & $281.9$ {\scriptsize[5.72]} \\
$16$ & $25.82$\textsuperscript{$\dagger$} {\scriptsize[4.87]} & $32.36$ {\scriptsize[5.02]} & $40.25$ {\scriptsize[5.17]} & $52.63$ {\scriptsize[5.30]} & $\mathbf{69.50}$ {\scriptsize[5.41]} & $91.69$ {\scriptsize[5.53]} & $124.4$ {\scriptsize[5.63]} \\
$32$ & $19.01$\textsuperscript{$\dagger$} {\scriptsize[4.88]} & $22.66$ {\scriptsize[5.03]} & $28.14$ {\scriptsize[5.16]} & $36.23$ {\scriptsize[5.27]} & $\mathbf{44.80}$ {\scriptsize[5.37]} & $61.95$ {\scriptsize[5.52]} & $83.03$ {\scriptsize[5.60]} \\
$64$ & $15.63$\textsuperscript{$\dagger$} {\scriptsize[4.87]} & $19.29$ {\scriptsize[5.05]} & $24.17$ {\scriptsize[5.16]} & $30.31$ {\scriptsize[5.26]} & $\mathbf{37.68}$ {\scriptsize[5.36]} & $49.26$ {\scriptsize[5.47]} & $69.53$ {\scriptsize[5.56]} \\
$128$ & $14.54$\textsuperscript{$\dagger$} {\scriptsize[4.88]} & $16.47$\textsuperscript{$\dagger$} {\scriptsize[4.97]} & $20.51$ {\scriptsize[5.11]} & $25.56$ {\scriptsize[5.22]} & $\mathbf{34.42}$ {\scriptsize[5.36]} & $44.01$ {\scriptsize[5.47]} & $60.19$ {\scriptsize[5.58]} \\
$256$ & $13.22$\textsuperscript{$\dagger$} {\scriptsize[4.74]} & $15.94$\textsuperscript{$\dagger$} {\scriptsize[4.94]} & $19.52$ {\scriptsize[5.04]} & $23.69$ {\scriptsize[5.17]} & $\mathbf{28.99}$ {\scriptsize[5.26]} & $40.14$ {\scriptsize[5.44]} & $54.84$ {\scriptsize[5.53]} \\
$512$ & $12.18$\textsuperscript{$\dagger$} {\scriptsize[4.72]} & $14.78$\textsuperscript{$\dagger$} {\scriptsize[4.90]} & $18.35$ {\scriptsize[5.04]} & $23.06$ {\scriptsize[5.14]} & $\mathbf{30.15}$ {\scriptsize[5.29]} & $40.63$ {\scriptsize[5.42]} & $54.27$ {\scriptsize[5.49]} \\
$1024$ & $11.71$\textsuperscript{$\dagger$} {\scriptsize[4.71]} & $14.73$\textsuperscript{$\dagger$} {\scriptsize[4.93]} & $19.40$ {\scriptsize[5.09]} & $24.15$ {\scriptsize[5.20]} & $\mathbf{30.28}$ {\scriptsize[5.30]} & $41.02$ {\scriptsize[5.46]} & $55.21$ {\scriptsize[5.55]} \\
\midrule
\multicolumn{8}{l}{\textbf{fp32 categorical sampling.}} \\
$1$ & $463.8$\textsuperscript{$\dagger$} {\scriptsize[4.40]} & $761.6$\textsuperscript{$\dagger$} {\scriptsize[4.76]} & $\mathbf{1287}$ {\scriptsize[5.11]} & $2138$ {\scriptsize[5.44]} & $3289$ {\scriptsize[5.70]} & $4874$ {\scriptsize[5.92]} & $6941$ {\scriptsize[6.12]} \\
$2$ & $279.3$\textsuperscript{$\dagger$} {\scriptsize[4.22]} & $457.6$\textsuperscript{$\dagger$} {\scriptsize[4.66]} & $\mathbf{842.3}$ {\scriptsize[5.11]} & $1432$ {\scriptsize[5.46]} & $2388$ {\scriptsize[5.76]} & $3875$ {\scriptsize[6.02]} & $5987$ {\scriptsize[6.24]} \\
$4$ & $107.9$\textsuperscript{$\dagger$} {\scriptsize[4.41]} & $153.3$\textsuperscript{$\dagger$} {\scriptsize[4.79]} & $\mathbf{235.5}$ {\scriptsize[5.12]} & $357.1$ {\scriptsize[5.36]} & $510.3$ {\scriptsize[5.54]} & $792.1$ {\scriptsize[5.75]} & $1194$ {\scriptsize[5.90]} \\
$8$ & $44.99$\textsuperscript{$\dagger$} {\scriptsize[4.69]} & $57.98$\textsuperscript{$\dagger$} {\scriptsize[4.95]} & $\mathbf{74.63}$ {\scriptsize[5.13]} & $103.0$ {\scriptsize[5.31]} & $139.1$ {\scriptsize[5.42]} & $195.2$ {\scriptsize[5.58]} & $287.3$ {\scriptsize[5.72]} \\
$16$ & $26.24$\textsuperscript{$\dagger$} {\scriptsize[4.88]} & $32.68$ {\scriptsize[5.01]} & $\mathbf{42.68}$ {\scriptsize[5.17]} & $55.42$ {\scriptsize[5.29]} & $67.49$ {\scriptsize[5.40]} & $88.35$ {\scriptsize[5.51]} & $122.6$ {\scriptsize[5.63]} \\
$32$ & $18.51$\textsuperscript{$\dagger$} {\scriptsize[4.87]} & $21.89$ {\scriptsize[5.03]} & $\mathbf{26.75}$ {\scriptsize[5.14]} & $35.68$ {\scriptsize[5.26]} & $42.60$ {\scriptsize[5.38]} & $58.88$ {\scriptsize[5.48]} & $74.58$ {\scriptsize[5.58]} \\
$64$ & $14.54$\textsuperscript{$\dagger$} {\scriptsize[4.83]} & $18.02$\textsuperscript{$\dagger$} {\scriptsize[4.98]} & $\mathbf{21.66}$ {\scriptsize[5.07]} & $28.21$ {\scriptsize[5.22]} & $33.67$ {\scriptsize[5.33]} & $44.21$ {\scriptsize[5.43]} & $59.08$ {\scriptsize[5.54]} \\
$128$ & $13.05$\textsuperscript{$\dagger$} {\scriptsize[4.78]} & $16.60$\textsuperscript{$\dagger$} {\scriptsize[4.97]} & $\mathbf{19.92}$ {\scriptsize[5.09]} & $24.31$ {\scriptsize[5.20]} & $31.25$ {\scriptsize[5.31]} & $37.76$ {\scriptsize[5.38]} & $50.78$ {\scriptsize[5.50]} \\
$256$ & $12.12$\textsuperscript{$\dagger$} {\scriptsize[4.72]} & $14.17$\textsuperscript{$\dagger$} {\scriptsize[4.84]} & $\mathbf{17.22}$\textsuperscript{$\dagger$} {\scriptsize[4.99]} & $20.79$ {\scriptsize[5.10]} & $26.15$ {\scriptsize[5.23]} & $33.92$ {\scriptsize[5.34]} & $44.05$ {\scriptsize[5.45]} \\
$512$ & $10.97$\textsuperscript{$\dagger$} {\scriptsize[4.67]} & $13.25$\textsuperscript{$\dagger$} {\scriptsize[4.88]} & $\mathbf{16.31}$\textsuperscript{$\dagger$} {\scriptsize[4.99]} & $20.39$ {\scriptsize[5.13]} & $24.45$ {\scriptsize[5.22]} & $31.73$ {\scriptsize[5.33]} & $39.61$ {\scriptsize[5.43]} \\
$1024$ & $10.69$\textsuperscript{$\dagger$} {\scriptsize[4.57]} & $12.88$\textsuperscript{$\dagger$} {\scriptsize[4.76]} & $\mathbf{15.23}$\textsuperscript{$\dagger$} {\scriptsize[4.91]} & $17.46$ {\scriptsize[5.00]} & $20.96$ {\scriptsize[5.10]} & $26.68$ {\scriptsize[5.22]} & $34.57$ {\scriptsize[5.36]} \\
\bottomrule
\end{tabular}}

\medskip
    \caption{\textbf{DiDi-Instruct, reward-guided ancestral sampling.} GPT-2-Large gPPL [$H$], 40 samples per cell. Bold: standard $(1,\mathrm{fp64})$ and sharpened $(.9,\mathrm{fp32})$; $\dagger$: $H<5$.}
    \label{tab:didi_full_grid_guided}
\setlength{\tabcolsep}{2.5pt}
\resizebox{.97\linewidth}{!}{%
\begin{tabular}{rccccccc}
\toprule
$M$ & $\tau{=}0.8$ & $\tau{=}0.85$ & $\tau{=}0.9$ & $\tau{=}0.95$ & $\tau{=}1.0$ & $\tau{=}1.05$ & $\tau{=}1.10$ \\
\midrule
\multicolumn{8}{l}{\textbf{fp64 categorical sampling.}} \\
$1$ & $513.8$\textsuperscript{$\dagger$} {\scriptsize[4.46]} & $849.1$\textsuperscript{$\dagger$} {\scriptsize[4.82]} & $1478$ {\scriptsize[5.19]} & $2357$ {\scriptsize[5.50]} & $\mathbf{3474}$ {\scriptsize[5.73]} & $5343$ {\scriptsize[5.97]} & $7516$ {\scriptsize[6.16]} \\
$2$ & $294.3$\textsuperscript{$\dagger$} {\scriptsize[4.28]} & $508.2$\textsuperscript{$\dagger$} {\scriptsize[4.73]} & $918.0$ {\scriptsize[5.17]} & $1607$ {\scriptsize[5.54]} & $\mathbf{2550}$ {\scriptsize[5.81]} & $4074$ {\scriptsize[6.07]} & $6249$ {\scriptsize[6.27]} \\
$4$ & $109.9$\textsuperscript{$\dagger$} {\scriptsize[4.41]} & $154.2$\textsuperscript{$\dagger$} {\scriptsize[4.80]} & $244.4$ {\scriptsize[5.12]} & $368.7$ {\scriptsize[5.37]} & $\mathbf{523.6}$ {\scriptsize[5.56]} & $771.5$ {\scriptsize[5.74]} & $1261$ {\scriptsize[5.93]} \\
$8$ & $43.88$\textsuperscript{$\dagger$} {\scriptsize[4.64]} & $61.45$\textsuperscript{$\dagger$} {\scriptsize[4.90]} & $77.23$ {\scriptsize[5.15]} & $103.7$ {\scriptsize[5.31]} & $\mathbf{147.9}$ {\scriptsize[5.45]} & $196.8$ {\scriptsize[5.56]} & $295.1$ {\scriptsize[5.72]} \\
$16$ & $24.59$\textsuperscript{$\dagger$} {\scriptsize[4.82]} & $31.50$\textsuperscript{$\dagger$} {\scriptsize[4.99]} & $40.95$ {\scriptsize[5.17]} & $52.10$ {\scriptsize[5.29]} & $\mathbf{71.56}$ {\scriptsize[5.43]} & $90.89$ {\scriptsize[5.52]} & $119.8$ {\scriptsize[5.62]} \\
$32$ & $20.04$\textsuperscript{$\dagger$} {\scriptsize[4.88]} & $22.88$ {\scriptsize[5.02]} & $27.64$ {\scriptsize[5.13]} & $36.92$ {\scriptsize[5.28]} & $\mathbf{46.76}$ {\scriptsize[5.39]} & $64.35$ {\scriptsize[5.50]} & $83.71$ {\scriptsize[5.59]} \\
$64$ & $15.24$\textsuperscript{$\dagger$} {\scriptsize[4.88]} & $18.76$ {\scriptsize[5.01]} & $23.41$ {\scriptsize[5.15]} & $27.85$ {\scriptsize[5.23]} & $\mathbf{37.28}$ {\scriptsize[5.37]} & $50.21$ {\scriptsize[5.46]} & $64.89$ {\scriptsize[5.55]} \\
$128$ & $14.20$\textsuperscript{$\dagger$} {\scriptsize[4.83]} & $16.85$\textsuperscript{$\dagger$} {\scriptsize[4.98]} & $19.73$ {\scriptsize[5.10]} & $25.19$ {\scriptsize[5.22]} & $\mathbf{34.51}$ {\scriptsize[5.35]} & $42.10$ {\scriptsize[5.43]} & $58.36$ {\scriptsize[5.55]} \\
$256$ & $13.42$\textsuperscript{$\dagger$} {\scriptsize[4.85]} & $15.39$\textsuperscript{$\dagger$} {\scriptsize[4.92]} & $18.42$ {\scriptsize[5.04]} & $23.93$ {\scriptsize[5.21]} & $\mathbf{30.93}$ {\scriptsize[5.33]} & $40.56$ {\scriptsize[5.43]} & $54.13$ {\scriptsize[5.55]} \\
$512$ & $11.92$\textsuperscript{$\dagger$} {\scriptsize[4.73]} & $15.25$\textsuperscript{$\dagger$} {\scriptsize[4.93]} & $18.60$ {\scriptsize[5.07]} & $24.22$ {\scriptsize[5.22]} & $\mathbf{30.82}$ {\scriptsize[5.33]} & $41.03$ {\scriptsize[5.44]} & $53.36$ {\scriptsize[5.53]} \\
$1024$ & $11.84$\textsuperscript{$\dagger$} {\scriptsize[4.72]} & $14.62$\textsuperscript{$\dagger$} {\scriptsize[4.89]} & $17.71$ {\scriptsize[5.05]} & $23.05$ {\scriptsize[5.20]} & $\mathbf{30.06}$ {\scriptsize[5.31]} & $39.70$ {\scriptsize[5.43]} & $58.32$ {\scriptsize[5.57]} \\
\midrule
\multicolumn{8}{l}{\textbf{fp32 categorical sampling.}} \\
$1$ & $497.6$\textsuperscript{$\dagger$} {\scriptsize[4.47]} & $821.1$\textsuperscript{$\dagger$} {\scriptsize[4.82]} & $\mathbf{1374}$ {\scriptsize[5.20]} & $2269$ {\scriptsize[5.51]} & $3438$ {\scriptsize[5.76]} & $5062$ {\scriptsize[5.98]} & $7214$ {\scriptsize[6.17]} \\
$2$ & $300.2$\textsuperscript{$\dagger$} {\scriptsize[4.29]} & $501.9$\textsuperscript{$\dagger$} {\scriptsize[4.73]} & $\mathbf{876.5}$ {\scriptsize[5.16]} & $1526$ {\scriptsize[5.51]} & $2448$ {\scriptsize[5.80]} & $4008$ {\scriptsize[6.05]} & $5990$ {\scriptsize[6.26]} \\
$4$ & $105.1$\textsuperscript{$\dagger$} {\scriptsize[4.44]} & $156.7$\textsuperscript{$\dagger$} {\scriptsize[4.84]} & $\mathbf{220.1}$ {\scriptsize[5.12]} & $335.0$ {\scriptsize[5.35]} & $483.0$ {\scriptsize[5.53]} & $756.6$ {\scriptsize[5.73]} & $1107$ {\scriptsize[5.89]} \\
$8$ & $43.88$\textsuperscript{$\dagger$} {\scriptsize[4.75]} & $57.22$\textsuperscript{$\dagger$} {\scriptsize[4.96]} & $\mathbf{76.09}$ {\scriptsize[5.14]} & $102.5$ {\scriptsize[5.32]} & $141.4$ {\scriptsize[5.43]} & $195.5$ {\scriptsize[5.58]} & $278.8$ {\scriptsize[5.71]} \\
$16$ & $24.93$\textsuperscript{$\dagger$} {\scriptsize[4.85]} & $31.53$ {\scriptsize[5.02]} & $\mathbf{40.66}$ {\scriptsize[5.17]} & $51.59$ {\scriptsize[5.28]} & $67.30$ {\scriptsize[5.39]} & $88.09$ {\scriptsize[5.52]} & $123.5$ {\scriptsize[5.62]} \\
$32$ & $17.77$\textsuperscript{$\dagger$} {\scriptsize[4.86]} & $21.54$\textsuperscript{$\dagger$} {\scriptsize[4.99]} & $\mathbf{26.73}$ {\scriptsize[5.13]} & $33.99$ {\scriptsize[5.25]} & $42.33$ {\scriptsize[5.37]} & $56.25$ {\scriptsize[5.48]} & $76.96$ {\scriptsize[5.59]} \\
$64$ & $14.55$\textsuperscript{$\dagger$} {\scriptsize[4.81]} & $16.91$\textsuperscript{$\dagger$} {\scriptsize[4.92]} & $\mathbf{20.74}$ {\scriptsize[5.07]} & $27.05$ {\scriptsize[5.21]} & $36.44$ {\scriptsize[5.35]} & $44.32$ {\scriptsize[5.45]} & $59.53$ {\scriptsize[5.56]} \\
$128$ & $12.99$\textsuperscript{$\dagger$} {\scriptsize[4.77]} & $15.72$\textsuperscript{$\dagger$} {\scriptsize[4.92]} & $\mathbf{18.91}$ {\scriptsize[5.05]} & $24.35$ {\scriptsize[5.20]} & $31.04$ {\scriptsize[5.31]} & $38.68$ {\scriptsize[5.43]} & $50.67$ {\scriptsize[5.52]} \\
$256$ & $11.91$\textsuperscript{$\dagger$} {\scriptsize[4.70]} & $14.78$\textsuperscript{$\dagger$} {\scriptsize[4.92]} & $\mathbf{17.04}$ {\scriptsize[5.03]} & $20.12$ {\scriptsize[5.13]} & $27.18$ {\scriptsize[5.27]} & $34.44$ {\scriptsize[5.37]} & $48.18$ {\scriptsize[5.49]} \\
$512$ & $10.98$\textsuperscript{$\dagger$} {\scriptsize[4.62]} & $13.21$\textsuperscript{$\dagger$} {\scriptsize[4.81]} & $\mathbf{15.88}$\textsuperscript{$\dagger$} {\scriptsize[4.97]} & $18.93$ {\scriptsize[5.08]} & $23.29$ {\scriptsize[5.20]} & $29.85$ {\scriptsize[5.33]} & $39.01$ {\scriptsize[5.43]} \\
$1024$ & $11.66$\textsuperscript{$\dagger$} {\scriptsize[4.74]} & $13.29$\textsuperscript{$\dagger$} {\scriptsize[4.83]} & $\mathbf{15.08}$\textsuperscript{$\dagger$} {\scriptsize[4.95]} & $17.49$ {\scriptsize[5.05]} & $21.28$ {\scriptsize[5.14]} & $26.19$ {\scriptsize[5.27]} & $33.09$ {\scriptsize[5.37]} \\
\bottomrule
\end{tabular}}
\end{table}

\clearpage
\subsection{The sharpened sampler is also memory and compute efficient}
\label{app:sampling_cost}

\begin{figure}[t]
    \centering
    \begin{subfigure}[t]{0.48\linewidth}
        \centering
        \includegraphics[width=\linewidth]{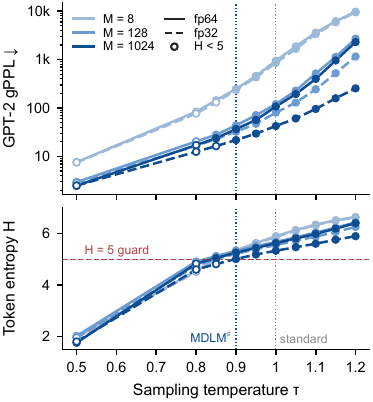}
        \caption{\(\tau\)-sweep of the public checkpoint at \(M\in\{8,128,1024\}\). gPPL falls under sharpening (top), while token entropy eventually crosses the \(H{\ge}5\) guard (bottom); filled markers pass the guard and open markers fail it. Dotted lines mark the standard \(\tau=1\) and the \sharpm{} \(\tau=.9\).}
        \label{fig:tau_frontier}
    \end{subfigure}
    \hfill
    \begin{subfigure}[t]{0.48\linewidth}
        \centering
        \includegraphics[width=\linewidth]{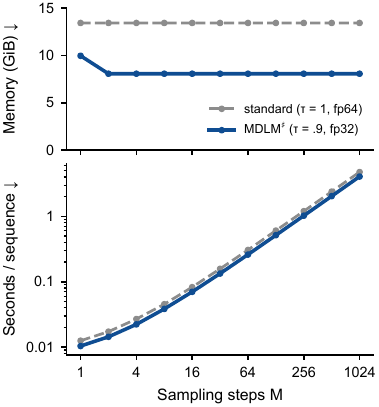}
        \caption{Matching-configuration cost of \sharpm{} \((\tau{=}.9,\mathrm{fp32})\) and the standard \((\tau{=}1,\mathrm{fp64})\) sampler: peak allocated memory (top) and amortized wall-clock time (bottom).}
        \label{fig:cost_main}
    \end{subfigure}
    \caption{\textbf{The sampler exposes an empirical quality--diversity--cost frontier.} \emph{Left:} gPPL and generated-sequence \(H\) show why neither metric alone suffices. \emph{Right:} cost is measured on the same H100 protocol for the two displayed operating points.}
    \label{fig:frontier_and_cost}
\end{figure}

\Cref{fig:cost_main} is a matching-configuration comparison of \sharpm{} \((\tau{=}.9,\mathrm{fp32})\) against the standard \((\tau{=}1,\mathrm{fp64})\) sampler on MDLM-Small for \(M=1,\ldots,1024\). At \(M=1024\), \sharpm{} uses \(8.069\) GiB peak allocated memory versus \(13.439\) GiB and takes \(4.065\) versus \(4.764\) amortized wall-seconds per sequence at batch size \(8\): \(40.0\%\) lower peak allocation and \(17.2\%\) higher throughput. Each point uses the H100 protocol above. This run includes the extra log-softmax path invoked by \(\tau\ne1\), so the comparison exactly matches the operating points shown in the quality figures.

\subsection{The unchanged baseline rivals its distilled successors}
\label{app:cross_model_entropy}

\paragraph{Full cross-model comparison.}
\Cref{tab:final_model_compare} gives the full comparison behind \Cref{fig:overview}a, in gPPL with within-sequence entropy in brackets.

\begin{table}[t]
\centering
\caption{\textbf{GPT-2-Large gPPL [within-sequence entropy] in one consolidated 40-sample stress test.} \(M\) is configured reverse updates, not NFE; each official sampler uses its terminal-noise-removal convention. Sharpened rows fix \((\tau{=}.9,\mathrm{fp32})\); all other rows use the standard \((\tau{=}1,\mathrm{fp64})\) sampler. DUO-DCD is the distilled DUO model. \(\dagger\): \(H<5\). MDLM uses the public checkpoint; DiDi uses the EMA student and reward-guided sampler.}
\label{tab:final_model_compare}
\small
\setlength{\tabcolsep}{4pt}
\resizebox{\linewidth}{!}{%
\begin{tabular}{lcccccc}
\toprule
Model & $M{=}4$ & $M{=}16$ & $M{=}64$ & $M{=}128$ & $M{=}256$ & $M{=}1024$ \\
\midrule
MDLM-Small & 1948.8 {\scriptsize[5.95]} & 344.2 {\scriptsize[5.79]} & 145.4 {\scriptsize[5.71]} & 117.7 {\scriptsize[5.66]} & 109.9 {\scriptsize[5.64]} & 103.8 {\scriptsize[5.64]} \\
\textbf{\sharpm{} (same checkpoint)} & \textbf{519.4} {\scriptsize[5.29]} & \textbf{103.6} {\scriptsize[5.36]} & \textbf{44.5} {\scriptsize[5.30]} & \textbf{33.3} {\scriptsize[5.23]} & \textbf{30.1} {\scriptsize[5.15]} & \textbf{21.4}\textsuperscript{$\dagger$} {\scriptsize[4.98]} \\
DUO & 518.3 {\scriptsize[5.55]} & 126.2 {\scriptsize[5.59]} & 86.2 {\scriptsize[5.60]} & 86.2 {\scriptsize[5.57]} & 70.1 {\scriptsize[5.54]} & 85.6 {\scriptsize[5.57]} \\
DUO-DCD & 314.9 {\scriptsize[5.51]} & 83.1 {\scriptsize[5.55]} & 66.0 {\scriptsize[5.52]} & 62.0 {\scriptsize[5.50]} & 51.2 {\scriptsize[5.46]} & 65.7 {\scriptsize[5.47]} \\
SEDD & 1863.3 {\scriptsize[5.91]} & 385.7 {\scriptsize[5.81]} & 146.5 {\scriptsize[5.69]} & 131.9 {\scriptsize[5.66]} & 100.1 {\scriptsize[5.62]} & 114.0 {\scriptsize[5.65]} \\
DiDi-Instruct & 523.6 {\scriptsize[5.56]} & 71.6 {\scriptsize[5.43]} & 37.3 {\scriptsize[5.37]} & 34.5 {\scriptsize[5.35]} & 30.9 {\scriptsize[5.33]} & 30.1 {\scriptsize[5.31]} \\
\textbf{DiDi-Instruct, sharpened} & \textbf{220.1} {\scriptsize[5.12]} & \textbf{40.7} {\scriptsize[5.17]} & \textbf{20.7} {\scriptsize[5.07]} & \textbf{18.9} {\scriptsize[5.05]} & \textbf{17.0} {\scriptsize[5.03]} & \textbf{15.1}\textsuperscript{$\dagger$} {\scriptsize[4.95]} \\
\midrule
GPT-2 generator & \multicolumn{6}{c}{$31.3$ {\scriptsize[5.50]} \quad (update-independent reference)} \\
\bottomrule
\end{tabular}}
\vspace{-4pt}
\end{table}

\begin{figure}[ht]
    \centering
    \includegraphics[width=0.93\linewidth]{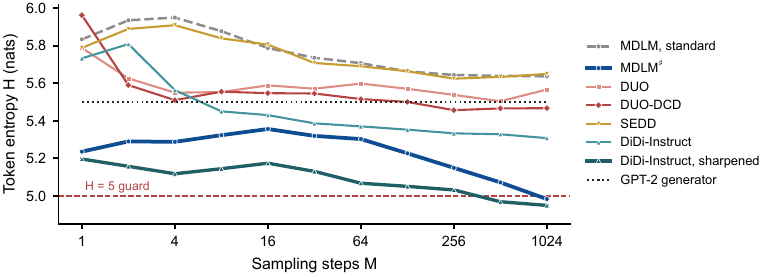}
    \caption{\textbf{Sharpened samplers stay near the \(H{=}5\) guard.} Within-sequence entropy versus configured updates in the cross-model stress test. The dashed line marks \(H=5\) and the dotted line the autoregressive GPT-2 generator; \sharpm{} and sharpened DiDi-Instruct cross the guard slightly at the largest budgets.}
    \label{fig:cross_model_entropy_appendix}
\end{figure}

\paragraph{Entropy guard diagnostic.}
\Cref{fig:cross_model_entropy_appendix} plots within-sequence entropy versus configured \(M\) for the cross-model stress test. \sharpm{} remains above the guard through \(M=256\) and is just below it at \(M=1024\) (\(H=4.98\)); sharpened DiDi-Instruct remains above it through \(M=256\) and also crosses slightly at \(M\ge512\) (\(H\approx4.95\) at \(M=1024\)). A \(\tau=.95\) DiDi variant passes the guard throughout and still improves substantially over the standard sampler (\Cref{tab:didi_full_grid_guided}). We mark the crossings rather than treating these diagnostics as a certificate of across-sample diversity or distributional correctness.

\subsection{Sharpening worsens the schedule-matched posterior loss}
\label{app:certificate_measurement}

We measure the per-position forced-mask objective \(\mathcal J_{M+1}/L\) from \Cref{eq:forced_mask_estimators} on 512 paired held-out sequences for the public, early, and late MDLM-Small checkpoints. We load EMA weights and use the idealized deployed reveal masses: \(M\) masses \((1-10^{-5})/M\) plus terminal cleanup mass \(10^{-5}\), hence \(M+1\) reveal blocks. Each sequence shares its corruption across temperatures; 95\% intervals use a paired sequence bootstrap with 10,000 replicates. A separate 32-level, 128-sequence profile provides the early/late shape diagnostic.

\begin{table}[ht]
\centering
\caption{\textbf{Sharpening worsens the schedule-matched posterior loss.} Entries are \(\{\mathcal J_{M+1}(\tau{=}.9)-\mathcal J_{M+1}(\tau{=}1)\}/L\) in per-position nats with paired 95\% intervals. Positive values mean worse teacher-forced cross-entropy. Every interval excludes zero; the temperature-grid optimum is \(1\) or \(1.05\).}
\label{tab:certificate_measurement}
\scriptsize
\setlength{\tabcolsep}{3.5pt}
\resizebox{\linewidth}{!}{%
\begin{tabular}{lccccc}
\toprule
Checkpoint & $M{=}2$ & $M{=}8$ & $M{=}32$ & $M{=}128$ & $M{=}1024$ \\
\midrule
Public & $+.038\ [.011,.064]$ & $+.052\ [.029,.076]$ & $+.045\ [.021,.068]$ & $+.051\ [.028,.073]$ & $+.036\ [.014,.058]$ \\
Early  & $+.047\ [.021,.075]$ & $+.055\ [.030,.081]$ & $+.041\ [.016,.068]$ & $+.068\ [.042,.094]$ & $+.037\ [.012,.061]$ \\
Late   & $+.041\ [.014,.067]$ & $+.056\ [.033,.081]$ & $+.048\ [.024,.072]$ & $+.055\ [.031,.078]$ & $+.043\ [.020,.066]$ \\
\bottomrule
\end{tabular}}
\end{table}

At \(\tau=1\), late-minus-early differences are \(-0.426,-0.546,-0.589,-0.617,-0.512\) nats at the five displayed budgets, with all paired intervals below zero. On the profile measurement, late is lower at every nontrivial clean-probability level; its left-grid mean is lower by \(0.674\) nats (95\% CI \(0.629\)--\(0.722\)). Thus neither the output improvement from \(\tau=.9\) nor the observed early/late gPPL reversal is explained by a lower posterior loss.

\clearpage
\subsection{Large diffusion LMs already sit at their sharp limit}
\label{app:large_models}

We evaluate LLaDA-8B-Instruct~\citep{nie2025llada} and Dream-v0-Base-7B~\citep{dream2025} on 256 GSM8K~\citep{cobbe2021gsm8k} examples and all 164 HumanEval~\citep{chen2021humaneval} problems, using the Fast-dLLM protocol~\citep{wu2025fastdllm}: fp32 sampling, length 256, block size 32. We sweep \(\tau\in\{0,.25,.5,.75,1,1.5,2\}\), with one greedy run and three explicit seeds at positive temperatures. HumanEval outputs are sanitized into complete programs before functional evaluation; the stock continuation scorer is inappropriate for these outputs.

\begin{figure}[ht]
    \centering
    \includegraphics[width=0.85\linewidth]{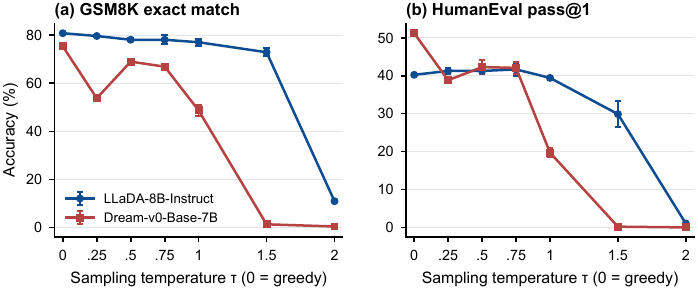}
    \caption{\textbf{For large diffusion LMs, sharpened sampler with lower temperature improves downstream performance.} LLaDA-8B-Instruct and Dream-v0-Base-7B on (a) GSM8K and (b) HumanEval. Greedy (temperature=0) is the default setting in official codebases.}
    \label{fig:large_model_temperature_stress}
\end{figure}

Greedy is best for both models on GSM8K and for Dream on HumanEval. LLaDA's HumanEval fluctuations through \(\tau=1\) are small relative to seed uncertainty; both models fail at sufficiently high temperature (\Cref{fig:large_model_temperature_stress}). These strong greedy defaults delimit the positive small-model finding: sharpening a positive-temperature baseline is not evidence that instruction-tuned defaults need changing.

\begin{figure}[ht]
    \centering
    \includegraphics[width=0.8\linewidth]{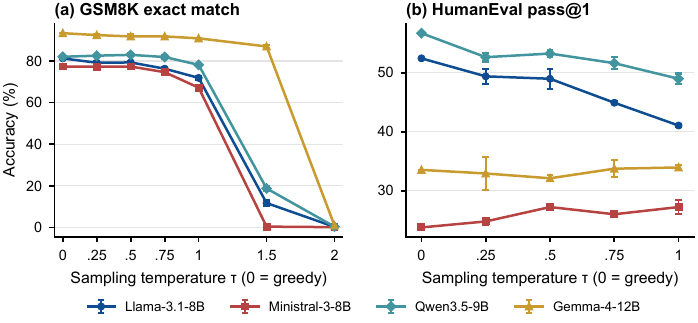}
    \caption{\textbf{Autoregressive control.} Llama-3.1-8B-Instruct, Ministral-3-8B-Instruct, Qwen3.5-9B, and Gemma-4-12B-it on (a) GSM8K, through \(\tau=2\), and (b) HumanEval, through \(\tau=1\), after complete-program extraction. Greedy runs once, while stochastic markers show three-seed means \(\pm1\) seed SEM.}
    \label{fig:autoregressive_temperature_stress}
\end{figure}

\paragraph{Autoregressive control.}
Four autoregressive checkpoints use the same task, length cap, and seed protocol, with unrestricted categorical sampling above \(\tau=0\) and thinking disabled for Qwen and Gemma (\Cref{fig:autoregressive_temperature_stress}). Their moderate-temperature responses also vary by model and task; high-temperature failure is not peculiar to diffusion. The diffusion models' greedy HumanEval scores (\(40.2\%,51.2\%\)) lie within the autoregressive range (\(23.8\%\)--\(56.7\%\)), but different checkpoints and post-training preclude a controlled model ranking.

%% file: sections/baseline_grids.tex
\subsection{The four-corner audit holds across checkpoints and budgets}
\label{app:baseline_grids}
\Cref{tab:headline_full} extends \Cref{tab:headline_main} to every budget and to our two MDLM-Small checkpoints.
\begin{table}[!ht]
\centering\small
\setlength{\tabcolsep}{4pt}
\caption{\textbf{A stronger unchanged baseline across checkpoints and budgets.} Three MDLM-Small checkpoints, 128 unconditional OpenWebText samples per cell; entries are GPT-2 gPPL with unigram entropy $H$ in brackets. The first four columns of settings form the four-corner audit; the last shows additional sharpening. \underline{Underlined}: standard sampler. \textbf{Bold}: best of the four corners based on both gPPL and $H$. $\dagger$: $H<5$. The standard sampler has the largest gPPL in every listed checkpoint--budget cell; $(\tau=.9,\mathrm{fp32})$ is best among the four corners for every $M\ge2$. At $M=1024$ it improves public-checkpoint gPPL by $4.9\times$ while passing the guard. $M$ counts configured updates, not NFE.}
\label{tab:headline_full}
\resizebox{\linewidth}{!}{%
\begin{tabular}{rccccc}
\toprule
$M$ & $(1,\mathrm{fp64})$ & $(1,\mathrm{fp32})$ & $(.9,\mathrm{fp64})$ & $(.9,\mathrm{fp32})$ & $(.85,\mathrm{fp32})$ \\
\midrule
\multicolumn{6}{l}{\textbf{(a) Public MDLM-Small \citep{sahoo2024simple} (the canonical baseline).}} \\
$1$ & $\underline{5043.3}$ {\small[5.83]} & $5001.7$ {\small[5.83]} & $\mathbf{2091.2}$ {\small[5.27]} & $2148.8$ {\small[5.26]} & $1306.6^{\dagger}$ {\small[4.91]} \\
$2$ & $\underline{3893.6}$ {\small[5.93]} & $3725.3$ {\small[5.93]} & $1293.3$ {\small[5.29]} & $\mathbf{1224.3}$ {\small[5.27]} & $675.8^{\dagger}$ {\small[4.84]} \\
$8$ & $\underline{945.1}$ {\small[5.89]} & $853.9$ {\small[5.87]} & $247.2$ {\small[5.34]} & $\mathbf{235.5}$ {\small[5.31]} & $131.6^{\dagger}$ {\small[4.98]} \\
$32$ & $\underline{196.8}$ {\small[5.73]} & $158.7$ {\small[5.70]} & $66.0$ {\small[5.35]} & $\mathbf{58.0}$ {\small[5.34]} & $40.0$ {\small[5.13]} \\
$128$ & $\underline{120.2}$ {\small[5.67]} & $80.2$ {\small[5.55]} & $43.7$ {\small[5.33]} & $\mathbf{32.3}$ {\small[5.21]} & $22.9$ {\small[5.01]} \\
$512$ & $\underline{105.9}$ {\small[5.64]} & $50.0$ {\small[5.40]} & $38.4$ {\small[5.28]} & $\mathbf{24.7}$ {\small[5.06]} & $18.3^{\dagger}$ {\small[4.88]} \\
$1024$ & $\underline{107.1}$ {\small[5.63]} & $42.5$ {\small[5.33]} & $36.6$ {\small[5.26]} & $\mathbf{21.8}$ {\small[5.01]} & $16.3^{\dagger}$ {\small[4.82]} \\
$2048$ & $\underline{106.5}$ {\small[5.63]} & $32.3$ {\small[5.18]} & $38.6$ {\small[5.28]} & $\mathbf{18.8}^{\dagger}$ {\small[4.89]} & $14.4^{\dagger}$ {\small[4.66]} \\
\midrule
\multicolumn{6}{l}{\textbf{(b) MDLM-Small late (this work).}} \\
$1$ & $\underline{4779.7}$ {\small[5.82]} & $4732.8$ {\small[5.82]} & $\mathbf{1992.2}$ {\small[5.25]} & $2044.9$ {\small[5.25]} & $1248.8^{\dagger}$ {\small[4.90]} \\
$2$ & $\underline{3634.1}$ {\small[5.93]} & $3513.4$ {\small[5.92]} & $1280.5$ {\small[5.31]} & $\mathbf{1219.4}$ {\small[5.29]} & $704.9^{\dagger}$ {\small[4.90]} \\
$8$ & $\underline{795.6}$ {\small[5.88]} & $725.3$ {\small[5.84]} & $236.4$ {\small[5.35]} & $\mathbf{231.8}$ {\small[5.32]} & $132.1$ {\small[5.00]} \\
$32$ & $\underline{194.5}$ {\small[5.72]} & $151.2$ {\small[5.68]} & $64.7$ {\small[5.35]} & $\mathbf{60.9}$ {\small[5.34]} & $40.7$ {\small[5.14]} \\
$128$ & $\underline{121.4}$ {\small[5.67]} & $80.1$ {\small[5.55]} & $44.3$ {\small[5.31]} & $\mathbf{33.8}$ {\small[5.20]} & $24.0$ {\small[5.00]} \\
$512$ & $\underline{111.9}$ {\small[5.64]} & $50.1$ {\small[5.38]} & $38.0$ {\small[5.26]} & $\mathbf{24.7}$ {\small[5.05]} & $18.0^{\dagger}$ {\small[4.84]} \\
$1024$ & $\underline{108.2}$ {\small[5.63]} & $41.9$ {\small[5.33]} & $38.6$ {\small[5.26]} & $\mathbf{22.5}$ {\small[5.03]} & $16.7^{\dagger}$ {\small[4.82]} \\
$2048$ & $\underline{107.2}$ {\small[5.62]} & $34.5$ {\small[5.19]} & $41.0$ {\small[5.27]} & $\mathbf{19.2}^{\dagger}$ {\small[4.87]} & $14.7^{\dagger}$ {\small[4.70]} \\
\midrule
\multicolumn{6}{l}{\textbf{(c) MDLM-Small early (this work).}} \\
$1$ & $\underline{4941.5}$ {\small[5.80]} & $4864.3$ {\small[5.80]} & $\mathbf{2006.8}$ {\small[5.22]} & $2085.2$ {\small[5.23]} & $1280.6^{\dagger}$ {\small[4.87]} \\
$2$ & $\underline{3278.0}$ {\small[5.89]} & $3104.6$ {\small[5.88]} & $1212.2$ {\small[5.31]} & $\mathbf{1151.1}$ {\small[5.29]} & $686.8^{\dagger}$ {\small[4.93]} \\
$8$ & $\underline{605.9}$ {\small[5.80]} & $551.5$ {\small[5.78]} & $235.2$ {\small[5.41]} & $\mathbf{213.7}$ {\small[5.39]} & $138.6$ {\small[5.15]} \\
$32$ & $\underline{251.2}$ {\small[5.70]} & $205.1$ {\small[5.65]} & $88.1$ {\small[5.34]} & $\mathbf{81.7}$ {\small[5.32]} & $53.6$ {\small[5.13]} \\
$128$ & $\underline{190.0}$ {\small[5.67]} & $120.6$ {\small[5.53]} & $70.2$ {\small[5.30]} & $\mathbf{50.4}$ {\small[5.17]} & $34.8^{\dagger}$ {\small[4.99]} \\
$512$ & $\underline{180.1}$ {\small[5.64]} & $75.3$ {\small[5.35]} & $60.1$ {\small[5.25]} & $\mathbf{36.2}$ {\small[5.02]} & $26.7^{\dagger}$ {\small[4.84]} \\
$1024$ & $\underline{166.1}$ {\small[5.62]} & $64.4$ {\small[5.30]} & $57.5$ {\small[5.23]} & $\mathbf{32.1}^{\dagger}$ {\small[4.95]} & $24.0^{\dagger}$ {\small[4.79]} \\
$2048$ & $\underline{172.6}$ {\small[5.63]} & $50.5$ {\small[5.18]} & $59.5$ {\small[5.26]} & $\mathbf{26.9}^{\dagger}$ {\small[4.86]} & $21.0^{\dagger}$ {\small[4.70]} \\
\bottomrule
\end{tabular}}
\end{table}

%% file: sections/group_eval_appendix.tex
\clearpage
\section{GroupEval: Protocol, Full Results, and Diagnostics}
\label{app:groupeval}
This appendix gives the proofs and protocol behind GroupEval, followed by the full judge-separated results. Sampler names follow \Cref{app:setup}, with every contrast comparing the standard $(1,\mathrm{fp64})$ with the sharpened $(.9,\mathrm{fp32})$ setting.

\subsection{Proof of \Cref{prop:collapse}}
\label{app:collapse_proof}
Let the output space $\mathcal X$ be finite and write $z(x)=(f(x),g(x))\in\mathbb R^2$. The map $Q\mapsto(\E_Qf,\E_Qg)$ is linear on the simplex $\Delta(\mathcal X)$, so its image is the convex polygon $Z=\operatorname{conv}\{z(x):x\in\mathcal X\}$. A Pareto-optimal value cannot be an interior point of $Z$, since a small move inside $Z$ would improve both coordinates. Every boundary point of a polygon lies on an edge $[z(x),z(x')]$ and is attained by a generator supported on $\{x,x'\}$. 

For the constrained problem, $\min\E_Qf$ subject to $\E_Qg\ge h$, $\sum_xQ(x)=1$, and $Q\ge0$ is a feasible, bounded linear program with two constraints besides nonnegativity whenever $h\le\max_xg(x)$; it therefore has an optimal basic solution with at most two nonzero coordinates. Token-pooled log-gPPL on fixed-length outputs and mean per-output unigram entropy are both of this form. 

Finally, if a generator places all its mass on two outputs, every group of $n=8$ samples contains at most two distinct texts. Identical texts share a mode under the rubric, so the group's mode entropy is at most $\log2$ and $\mathrm{\textsc{GED}}_g\le10\log2/\log8=10/3$.

\subsection{Elementary properties and limits of \textsc{GED}}
\label{app:ged_properties}
For a fixed group size $n\ge2$, let valid-mode counts be $n_1,\ldots,n_K$ with $V=\sum_kn_k\le n$. Multiplying entropy by the valid count rewrites the group score as
\begin{equation}
 \mathrm{\textsc{GED}}_g=\frac{10}{n\log n}
 \left[V\log V-\sum_k n_k\log n_k\right],
 \label{eq:ged_counts}
\end{equation}
with the continuous convention $0\log0=0$ and zero for an empty group. This expression also gives zero for a single valid sample or a single mode.

\paragraph{Range, invariances, and mode merging.}
The score is invariant to sample order and cluster labels. Since mode entropy lies between zero and $\log V$,
$0\le\mathrm{\textsc{GED}}_g\le10(V/n)(\log V/\log n)\le10$ for $V\ge2$.
Equality at ten holds exactly when $V=n$ and all $n$ modes are singletons. Merging counts $x,y>0$ replaces $x\log x+y\log y$ by $(x+y)\log(x+y)$, a larger quantity, so it cannot increase \textsc{GED}.

\paragraph{The validity gate does not reward removal at fixed assignments.}
Invalidating one sample in a mode of size $m\le V$ changes the bracket in \Cref{eq:ged_counts} by subtracting
\begin{equation}
 f(V)-f(m),\qquad f(x)=x\log x-(x-1)\log(x-1).
\end{equation}
For $x>1$, $f'(x)=\log(x/(x-1))>0$, and the endpoint is interpreted continuously. Thus $f(V)\ge f(m)$ and the score cannot increase. This proves the stated monotonicity under fixed assignments. A judge may recluster the remaining outputs when a group changes, so the property does not assert empirical robustness of the clustering procedure.

\paragraph{Local saturation is not support coverage.}
Suppose all outputs are valid and their true modes are independently uniform on $K\ge n$ modes. The probability that all group modes are distinct is
$\prod_{j=0}^{n-1}(1-j/K)$, at least $1-\binom n2/K$ by a collision union bound. On that event \textsc{GED} is ten. As $K$ grows, a fixed small group therefore almost always saturates, while observing a vanishing fraction of the total modes. This is why GroupEval reports ceiling rates and does not interpret high \textsc{GED} as recovery of the data distribution.

\paragraph{Why weighted \textsc{GED} ranges are not ordinary confidence intervals.}
Consider an all-valid, all-singleton group. The unweighted \textsc{GED} is ten. If its sample weights are replaced by a nonuniform Dirichlet draw $W$ and normalized within the group, its entropy becomes $H(W)<\log n$ almost surely. The resulting weighted score is strictly below ten even though the labels have not changed. A percentile range from those draws can therefore exclude the observed score systematically. We retain such ranges only as conditional weight sensitivity summaries in the released aggregates, not as calibrated intervals around plug-in \textsc{GED}. \textsc{GEQ}, being an average of fixed quality scores rather than an entropy functional, does not have this particular centering issue.

\subsection{Evaluation protocol}
\subsubsection{Rubric and deterministic scores}
\label{app:ge_rubric}
GroupEval v1.0 evaluates individual quality and local across-output diversity separately. The frozen prompt requests structured rubric decisions rather than a direct overall preference or an explanation of which system should win.

\paragraph{Quality.}
For sample $i$ in grouping context $r$, a catastrophic-invalidity gate marks empty, predominantly corrupted, catastrophically unreadable, or non-communicative text invalid. Weak but usable text remains valid. The four dimensions are readability/local fluency, coherence/continuity, substance/target naturalness, and non-degeneration/document structure. Each receives $s_{ird}\in\{0,1,2\}$, with all four set to zero for an invalid output. Thus
\begin{equation}
 q_{ir}=\frac54\sum_{d=1}^4s_{ird},\qquad
 \operatorname{\textsc{GEQ}}=\frac1S\sum_{i=1}^S\frac1R\sum_{r=1}^Rq_{ir}.
 \label{eq:groupeval-geq}
\end{equation}
Invalid outputs lower the mean rather than disappearing through filtering. The report also includes invalidity, severe-degeneration rates, dimension means, the 10th percentile, and bottom-quartile quality.

\paragraph{Diversity.}
Every valid sample receives one hard, group-local semantic--discourse mode. Modes reflect substantive content, purpose or genre, narrative setup, argument or perspective, or discourse structure. Paraphrases, entity substitutions within one template, formatting differences, and random corruption do not create useful modes. \textsc{GED} is computed by \Cref{eq:ged_main}; equivalently, \Cref{eq:ged_counts} expresses it directly in mode counts. 
The score is zero for fewer than two valid outputs or fewer than two modes. The validity gate is binary: a low-quality but valid output retains full mode mass, so \textsc{GEQ} must accompany \textsc{GED}. We report effective mode count, dominant-mode share, hybrid-assignment rate, clustering confidence, all-singleton rate, and \textsc{GED}-ceiling rate. The last two differ when some outputs are invalid. \Cref{app:ged_properties} proves the score's range and elementary invariances and explains local saturation.

\paragraph{Comparison and reuse.}
A common protocol makes comparisons interpretable across checkpoints, samplers, and budgets. Its identity includes the target-domain profile, generation conditions, judge version or available provenance, rubric, group size, preprocessing, grouping design, and scoring procedure. Each unconditional group contains outputs from one fixed generator setting. Validity and mode labels are inferred anew within each group; a mode label has no identity across groups. Comparisons keep these choices fixed and use the same output count where possible. Changing group size changes the collision probability and the $\log n$ normalization, so \textsc{GED} values at different group sizes are not interchangeable. The same judge name and group size alone do not make scores interchangeable, especially when a service alias can change. Judge series remain separate, and task-specific \textsc{GEQ} rubrics define different scales. The released prompts, schemas, grouping code, and deterministic scorer support reuse without imposing a universal quality--diversity threshold.

\subsubsection{Outputs, judges, and uncertainty}
\label{app:ge_default}
\paragraph{The shared unconditional configuration.}
Every main-text unconditional GroupEval number uses the following configuration, which we recommend as a common profile for unconditional text generators. This configuration takes into account practical limitations, especially compute costs.
\begin{itemize}
\setlength\itemsep{1pt}
\item \textbf{Groups.} $n=8$ outputs from one generator setting, randomly ordered. Each output pool is partitioned twice; the second partition is chosen from 2,000 deterministic candidates to minimize reuse of output pairs. Every output is judged twice, and none is filtered.
\item \textbf{Judge.} Open-weight \texttt{gpt-oss-120b} served with vLLM, with tools and web access disabled; the serving fingerprint is recorded. Generator identity, sampler, budget, and automatic scores are hidden, and candidate text is delimited as untrusted data.
\item \textbf{Rubric.} The frozen v1.0 web-text profile: an invalidity gate, four quality dimensions on $\{0,1,2\}$, and one hard semantic--discourse mode per valid output, returned in a validated JSON schema.
\item \textbf{Scores.} \textsc{GEQ} (\Cref{eq:groupeval-geq}) over all outputs; \textsc{GED} (\Cref{eq:ged_main}) averaged over groups; never combined.
\item \textbf{Uncertainty and diagnostics.} \textsc{GEQ} intervals from a 10,000-replicate partition-blocked Bayesian bootstrap; \textsc{GED} reported with invalidity, singleton, and ceiling rates.
\end{itemize}
Comparisons keep all of these fixed. Robustness judges (\texttt{gemma4:26b-a4b}, \texttt{qwen3:32b}, \texttt{gpt-5.6-luna}, \texttt{gpt-5.6-terra}) receive identical groups and are reported separately, never pooled.

\paragraph{Multiple judges across MDLM and DiDi.}
The MDLM study began with three open-weight judges (GPT-OSS, Gemma, Qwen); frontier Luna and Terra were then added for both MDLM and DiDi. Thus GPT-OSS, Luna, and Terra form the shared panel, with Gemma and Qwen providing additional MDLM sensitivity checks. Across the initial study's ten sampler--budget cells, pairwise Spearman correlations are $.90$--$.95$ for \textsc{GEQ} but $.48$--$.90$ for \textsc{GED}: quality rankings agree more closely than diversity rankings. These correlations include variation across budgets; the within-budget contrasts below test the sharpening gain directly.

\paragraph{Frozen outputs and grouping.}
Public MDLM-Small uses all 128 archived outputs in each of ten cells: two fixed settings at $M\in\{4,16,64,256,1024\}$. Groups of eight follow the two frozen partitions described above. Each judge evaluates 32 groups per cell, or 320 accepted judgments. The five MDLM judge series therefore contain 1,600 accepted judgments of the same output pools. Display orders are randomized independently. The archived runs record a global Lightning seed, not per-sample seeds; shared grouping indices do not establish sample-level common random numbers.

Each saved sequence has 1,024 nominal positions, including one nominal BOS. Preprocessing removes exactly one leading \texttt{<|endoftext|>} sentinel when present, preserves a failure to generate it, and renders every subsequent sentinel as \texttt{[DOCUMENT BOUNDARY]}. Nothing after a boundary is discarded. A cross-boundary topic change is not itself a coherence error; each document is assessed normally. The unconditional sweep's gPPL instead masks content after the first non-leading EOS. These columns are therefore not span-identical to GroupEval. The matched rescore below explicitly scores the full stream and retains first-document scoring as a sensitivity analysis.

\paragraph{Judge configuration and provenance.}
The dated MDLM series use \texttt{gpt-oss:120b} (July 25, 2026), \texttt{gemma4:26b-a4b} (July 26), \texttt{qwen3:32b} (July 27), and \texttt{gpt-5.6-luna} and \texttt{gpt-5.6-terra} (July 31). Each alias is a separate judge series, with no pooling or scale alignment.

Every accepted judgment passes local validation against the frozen schema, regardless of the provider's structured-output support. The archive records model metadata, timestamps, retries, and prompt/schema/profile and response hashes. The supplementary release contains aggregate scores, diagnostics, manifests, prompts, and analysis code, but not raw generations or individual judgments; exact reaggregation therefore requires the original archives. Mutable service aliases do not identify an immutable model snapshot.

\paragraph{Uncertainty.}
Unconditional contrasts use a 10,000-replicate partition-blocked Bayesian bootstrap~\citep{rubin1981bayesian}. Partition weights are shared across compared settings; sample weights are independent because the outputs are not sample-paired. Scores are recomputed within every weighted replicate. \textsc{GEQ} intervals condition on accepted judgments and the two frozen partitions: they do not include rejudging, new partitions, service drift, or new generation runs. For \textsc{GED}, unequal sample weights can systematically reduce entropy in singleton groups. Its weighted percentile ranges are therefore weight-sensitivity diagnostics, not confidence intervals centered on the plug-in \textsc{GED}. No diversity conclusion uses their exclusion of zero. The task-conditioned study uses a condition-cluster bootstrap instead.

\subsection{Unconditional generation}
\subsubsection{MDLM-Small: full judge-separated results}
\label{app:ge_mdlm}
\Cref{tab:groupeval-mdlm-all} brings all five judges into one comparison. At $M\ge16$, all twenty quality contrasts favor \sharpm{}, and thirteen conditional 95\% intervals exclude zero. Diversity rises in sixteen of twenty contrasts, despite lower unigram entropy in every pair. Differences use unrounded aggregates; the released CSVs retain all score and uncertainty diagnostics.

\begin{table}[!ht]
\centering\small
\caption{\textbf{MDLM improves in quality under all five judges at every primary budget.} \textsc{GEQ} and \textsc{GED} are standard $\to$ \sharpm{} on the same 128 outputs per cell. $\Delta$\textsc{GEQ} is sharpened minus standard, with conditional 95\% intervals; bold marks intervals excluding zero before rounding. All budgets, including the $M=4$ floor case, are retained.}
\label{tab:groupeval-mdlm-all}
\setlength{\tabcolsep}{10pt}
\begin{tabular}{rccc}
\toprule
Steps $M$ & \textsc{GEQ} $\uparrow$ & $\Delta$\textsc{GEQ} [95\% interval] & \textsc{GED} $\uparrow$ \\
\midrule
\multicolumn{4}{l}{\textbf{GPT-OSS (default)}}\\
% Source: group-eval/results/mdlm_pilot_gpt_oss_120b_20260725/pilot_{results,deltas}.csv
4 & $1.66\to1.53$ & $-0.12\ [-0.61,0.35]$ & $0.79\to1.11$ \\
16 & $3.85\to4.36$ & $+0.50\ [-0.08,1.08]$ & $6.00\to7.57$ \\
64 & $4.69\to5.00$ & $+0.31\ [-0.21,0.82]$ & $8.28\to8.79$ \\
256 & $4.92\to5.99$ & $\mathbf{+1.07\ [0.42,1.73]}$ & $8.67\to9.02$ \\
1024 & $5.15\to5.47$ & $+0.32\ [-0.20,0.85]$ & $8.07\to8.75$ \\
\midrule
\multicolumn{4}{l}{\textbf{Gemma}}\\
% Source: group-eval/results/mdlm_pilot_gemma4_26b_a4b_20260726/pilot_{results,deltas}.csv
4 & $2.93\to2.57$ & $-0.36\ [-0.76,0.05]$ & $6.76\to4.52$ \\
16 & $4.94\to5.20$ & $\mathbf{+0.26\ [0.00,0.55]}$ & $8.90\to9.24$ \\
64 & $5.77\to6.11$ & $\mathbf{+0.34\ [0.04,0.65]}$ & $9.86\to8.93$ \\
256 & $6.22\to6.42$ & $+0.21\ [-0.14,0.57]$ & $9.53\to9.14$ \\
1024 & $6.13\to6.43$ & $+0.29\ [-0.09,0.66]$ & $9.51\to9.27$ \\
\midrule
\multicolumn{4}{l}{\textbf{Qwen}}\\
% Source: group-eval/results/mdlm_pilot_qwen3_32b_20260727/pilot_{results,deltas}.csv
4 & $3.15\to4.72$ & $\mathbf{+1.57\ [0.87,2.27]}$ & $2.76\to4.82$ \\
16 & $5.24\to5.28$ & $+0.04\ [-0.33,0.39]$ & $7.28\to8.28$ \\
64 & $5.12\to5.85$ & $\mathbf{+0.74\ [0.36,1.11]}$ & $9.21\to9.70$ \\
256 & $5.47\to6.92$ & $\mathbf{+1.45\ [0.58,2.29]}$ & $9.12\to9.58$ \\
1024 & $5.77\to7.68$ & $\mathbf{+1.90\ [1.33,2.45]}$ & $9.53\to9.80$ \\
\midrule
\multicolumn{4}{l}{\textbf{Luna}}\\
% Source: group-eval/results/mdlm_pilot_gpt_5_6_luna_20260731/pilot_{results,deltas}.csv
4 & $1.52\to1.92$ & $+0.40\ [-0.43,1.19]$ & $1.64\to3.39$ \\
16 & $4.47\to5.37$ & $\mathbf{+0.90\ [0.34,1.49]}$ & $8.77\to9.70$ \\
64 & $5.47\to5.94$ & $\mathbf{+0.47\ [0.24,0.71]}$ & $9.86\to9.67$ \\
256 & $5.61\to6.06$ & $\mathbf{+0.45\ [0.20,0.71]}$ & $9.59\to9.69$ \\
1024 & $5.69\to6.22$ & $\mathbf{+0.53\ [0.30,0.79]}$ & $9.53\to9.70$ \\
\midrule
\multicolumn{4}{l}{\textbf{Terra}}\\
% Source: group-eval/results/mdlm_pilot_gpt_5_6_terra_20260731/pilot_{results,deltas}.csv
4 & $0.00\to0.00$ & $+0.00\ [0.00,0.00]$ & $0.00\to0.00$ \\
16 & $1.00\to3.44$ & $\mathbf{+2.45\ [1.55,3.30]}$ & $1.46\to5.60$ \\
64 & $4.72\to5.62$ & $\mathbf{+0.90\ [0.56,1.28]}$ & $8.03\to8.89$ \\
256 & $5.20\to5.74$ & $\mathbf{+0.54\ [0.24,0.87]}$ & $8.54\to9.61$ \\
1024 & $5.37\to5.65$ & $+0.28\ [-0.07,0.68]$ & $8.80\to9.54$ \\
\bottomrule
\end{tabular}
\end{table}

\paragraph{Failure cases and saturation.}
At $M=4$, GPT-OSS and Gemma favor the standard sampler in point estimate, Qwen and Luna favor \sharpm{}, and Terra rejects both. GPT-OSS marks $59.4\%$ of standard and $54.7\%$ of \sharpm{} sample contexts invalid at this budget. Lower gPPL alone does not rescue this regime.

Under GPT-OSS, mean effective-mode counts span $1.08$--$6.96$, all-singleton rates $9.4\%$--$62.5\%$, and \textsc{GED}-ceiling rates $0\%$--$50.0\%$ across all ten cells. At $M\ge16$, Luna's all-singleton and ceiling rates are substantially higher: $68.8\%$--$87.5\%$ and $65.6\%$--$87.5\%$, respectively; hybrid assignments reach $84.3\%$ and non-high confidence $87.5\%$. Terra's corresponding singleton and ceiling ranges are $18.8\%$--$75.0\%$ and $6.3\%$--$68.8\%$. These diagnostics limit the resolution of small \textsc{GED} differences, particularly under Luna. Every cell's failure, mode, and confidence diagnostics are retained in the supplementary CSVs.

\subsubsection{DiDi-Instruct: a judge-sensitive boundary case}
\label{app:ge_didi}
The reward-guided DiDi-Instruct EMA student~\citep{zheng2025didi} uses the same target profile, prompt, and scoring rules, but 40 outputs per cell. Two partitions into groups of eight yield ten judgments per cell, or 100 per judge. GPT-OSS, Luna, and Terra each score all ten cells. These are repeated judgments of fixed outputs, not independent generation replications. The standalone DiDi gPPL uses GPT-2-Large; it is not numerically comparable with the GPT-2 columns for the 128-output MDLM study.

\begin{table}[!ht]
\centering\small
\caption{\textbf{DiDi-Instruct's perplexity gain does not yield a judge-robust quality gain.} Same layout as \Cref{tab:groupeval-mdlm-all}, with 40 outputs per cell. GPT-OSS and Terra have no resolved quality gain; Luna favors the standard sampler at two primary budgets.}
\label{tab:groupeval-didi-all}
\setlength{\tabcolsep}{10pt}
\begin{tabular}{rccc}
\toprule
Steps $M$ & \textsc{GEQ} $\uparrow$ & $\Delta$\textsc{GEQ} [95\% interval] & \textsc{GED} $\uparrow$ \\
\midrule
\multicolumn{4}{l}{\textbf{GPT-OSS (default)}}\\
% Source: group-eval/results/didi_pilot_gpt_oss_120b_20260727/pilot_{results,deltas}.csv
4 & $1.59\to1.75$ & $+0.16\ [-0.68,1.00]$ & $1.81\to1.73$ \\
16 & $3.53\to3.72$ & $+0.19\ [-0.65,0.96]$ & $2.45\to4.20$ \\
64 & $4.48\to4.48$ & $+0.00\ [-0.75,0.70]$ & $4.86\to7.46$ \\
256 & $4.33\to5.03$ & $+0.70\ [-0.21,1.57]$ & $5.25\to7.63$ \\
1024 & $4.38\to5.08$ & $+0.70\ [-0.16,1.63]$ & $6.04\to7.51$ \\
\midrule
\multicolumn{4}{l}{\textbf{Luna}}\\
% Source: group-eval/results/didi_pilot_gpt_5_6_luna_20260731/pilot_{results,deltas}.csv
4 & $2.59\to2.62$ & $+0.03\ [-0.86,0.79]$ & $9.16\to9.82$ \\
16 & $4.86\to4.53$ & $\mathbf{-0.33\ [-0.68,-0.02]}$ & $8.43\to8.27$ \\
64 & $4.81\to4.50$ & $-0.31\ [-0.67,0.01]$ & $9.00\to8.15$ \\
256 & $4.86\to4.84$ & $-0.02\ [-0.29,0.23]$ & $9.19\to9.72$ \\
1024 & $4.98\to4.69$ & $\mathbf{-0.30\ [-0.59,-0.03]}$ & $9.72\to8.93$ \\
\midrule
\multicolumn{4}{l}{\textbf{Terra}}\\
% Source: group-eval/results/didi_pilot_gpt_5_6_terra_20260731/pilot_{results,deltas}.csv
4 & $0.00\to0.00$ & $+0.00\ [0.00,0.00]$ & $0.00\to0.00$ \\
16 & $2.53\to3.12$ & $+0.59\ [-0.10,1.26]$ & $1.62\to0.92$ \\
64 & $4.45\to4.81$ & $+0.36\ [-0.24,0.99]$ & $4.69\to6.66$ \\
256 & $4.50\to4.50$ & $+0.00\ [-0.49,0.52]$ & $5.14\to7.37$ \\
1024 & $4.58\to4.48$ & $-0.09\ [-0.57,0.42]$ & $7.59\to8.89$ \\
\bottomrule
\end{tabular}
\end{table}

GPT-OSS gives the sharpened sampler a higher or tied \textsc{GEQ} point estimate at every budget, but all its intervals include zero. Luna instead favors the standard sampler at all four primary budgets, with intervals below zero at $M\in\{16,1024\}$. Terra is mixed and all its intervals include zero. The DiDi quality improvement is therefore not judge-robust. \textsc{GED} directions are also mixed. Luna marks $76.3\%$--$100\%$ of valid DiDi assignments hybrid; $50\%$--$90\%$ of groups reach the \textsc{GED} ceiling and low clustering confidence reaches $90\%$. Its near-zero invalidity even at $M=4$ contrasts sharply with GPT-OSS and Terra. These observations expose rubric and mode-assignment sensitivity rather than justify selecting a preferred judge.

\subsubsection{Exact-output cross-generator comparison}
\label{app:ge_exact}
The cross-generator study evaluates the exact 40 outputs behind each MDLM and DiDi cell in \Cref{tab:final_model_compare}: two generators, two settings, and five budgets, totaling 800 outputs. A single globally interleaved, blinded GPT-OSS series provides 200 homogeneous-group judgments. Contrasts are matched by setting and budget and blocked by partition, not by sample index. 
% This is a separate, matched cohort; the standalone 128-output MDLM and 40-output DiDi series are not pooled into a cross-generator ranking.

Both GPT-2-Large and pretrained Qwen3-8B-Base~\citep{yang2025qwen3} rescore these same outputs with pinned evaluator revisions. The primary full-stream policy splits at every source document boundary, resets causal context, and pools token negative log-likelihoods and counts before exponentiation. First-document scoring, equal-output macro-gPPL, and bits per byte are sensitivity analyses.

\begin{table}[!ht]
  \centering
  \scriptsize
  \setlength{\tabcolsep}{2.8pt}
  \caption{\textbf{Likelihood and judged quality disagree on the same outputs.} gPPL columns are MDLM / DiDi ratios, with ratio$>1$ favoring DiDi. \textsc{GEQ}/\textsc{GED} entries are MDLM / DiDi, and $\Delta$\textsc{GEQ} is MDLM minus DiDi, with positive values favoring MDLM.}
  \label{tab:groupeval-exact40-common}
  \resizebox{\linewidth}{!}{%
  \begin{tabular}{rlccccc}
    \toprule
    \(M\) & Sampler
    & GPT-2-Large ratio
    & Qwen ratio
    & \textsc{GEQ}\(\uparrow\)
    & \(\Delta\)\textsc{GEQ} [95\% interval]
    & \textsc{GED}\(\uparrow\) \\
    \midrule
    16 & Std. & 4.58 & 4.69
      & 3.77 / 3.77 & 0.00 [-1.03, 1.02] & 6.88 / 4.27 \\
    16 & Sharp. & 2.33 & 2.71
      & 4.34 / 3.16 & 1.19 [0.41, 1.99] & 6.61 / 1.96 \\
    64 & Std. & 3.78 & 3.68
      & 4.30 / 4.09 & 0.20 [-0.85, 1.21] & 8.02 / 5.27 \\
    64 & Sharp. & 2.19 & 2.42
      & 4.64 / 4.23 & 0.41 [-0.26, 1.12] & 9.24 / 7.45 \\
    256 & Std. & 3.49 & 3.58
      & 5.70 / 4.20 & 1.50 [0.39, 2.65] & 8.70 / 6.59 \\
    256 & Sharp. & 1.76 & 1.95
      & 5.41 / 4.69 & 0.72 [-0.21, 1.64] & 9.27 / 6.23 \\
    1024 & Std. & 3.31 & 3.60
      & 5.09 / 4.33 & 0.77 [-0.25, 1.73] & 7.85 / 5.24 \\
    1024 & Sharp. & 1.48 & 1.56
      & 5.53 / 4.47 & 1.06 [-0.07, 2.22] & 9.30 / 7.08 \\
    \bottomrule
  \end{tabular}}
\end{table}

At $M\ge16$, both likelihood evaluators favor DiDi in all eight comparisons, with closely agreeing cell rankings across the 16 generator-setting-budget cells (Spearman $\rho=.982$). MDLM's \textsc{GEQ} is strictly higher in seven comparisons and tied in one; the intervals exclude zero for sharpened $M=16$ and standard $M=256$. Its \textsc{GED} is higher in all eight. First-document scoring retains only $42.97\%$ of source bytes but preserves every primary token-micro and bits-per-byte cross-model sign. Aggregation still matters: under equal-output macro-gPPL on first documents, Qwen favors MDLM in three of eight comparisons. Thus the stable primary conclusion is specific to the declared token-pooled metric, and may not extend to every aggregation rule.

Within each generator, likelihood is more informative about sampler selection. Excluding $M=4$, descriptive \textsc{GEQ}/log-gPPL Spearman correlations are $-.57$ and $-.74$ for MDLM under GPT-2-Large and Qwen, respectively, and $-.93$ for DiDi under either evaluator. Pooling generators weakens these to $-.20$ and $-.18$. This demonstrates a cross-generator disagreement between constructs, not that either scale is ground truth.

\subsection{Task-conditioned criterion check}
\label{app:ge_conditional}
We evaluate the greedy LLaDA-8B-Instruct and Dream-v0-Base-7B generations in \Cref{app:large_models}, using all 256 aligned GSM8K conditions and 164 aligned, sanitized HumanEval conditions. For each condition, GPT-OSS receives two randomized, model-blinded responses and scores readability, reasoning coherence, apparent task success, and non-degeneration. It sees no automatic labels, GSM8K answers, HumanEval hidden tests or canonical solutions, or aggregate results, and has no tools.

\begin{table}[!ht]
  \centering
  \scriptsize
  \setlength{\tabcolsep}{2.4pt}
  \caption{\textbf{Conditional \textsc{GEQ} tracks task success and the task-dependent model ordering.} L / D: LLaDA / Dream. Intervals: 10,000-replicate condition-cluster bootstrap, conditional on one GPT-OSS judgment per response pair. AUROC pools both models; the last column uses conditions with exactly one passing response (counts in parentheses). Task-specific c\textsc{GEQ} is not comparable with unconditional \textsc{GEQ}; \textsc{GED} is undefined with one response per model and condition.}
  \label{tab:groupeval-conditional-large}
  \resizebox{\linewidth}{!}{%
  \begin{tabular}{lccccc}
    \toprule
    Task & Automatic success L / D & c\textsc{GEQ} L / D
    & \(\Delta\)c\textsc{GEQ} D--L [95\% CI]
    & AUROC [95\% CI]
    & c\textsc{GEQ}(pass)--c\textsc{GEQ}(fail) [95\% CI] \\
    \midrule
    GSM8K & 80.9 / 75.4\% & 9.54 / 8.99
      & -0.55 [-0.77, -0.34] & 0.91 [0.87, 0.95]
      & 3.04 [2.54, 3.53] (56) \\
    HumanEval & 40.2 / 51.2\% & 8.13 / 8.40
      & 0.27 [-0.13, 0.65] & 0.96 [0.94, 0.98]
      & 3.07 [2.53, 3.66] (42) \\
    \bottomrule
  \end{tabular}}
\end{table}

c\textsc{GEQ} matches the automatic-success ordering on both tasks, although the aggregate HumanEval difference remains statistically unresolved. Pooled AUROC is $.91$ on GSM8K and $.96$ on HumanEval; all four task-by-model AUROCs exceed $.85$. On the 56 GSM8K and 42 HumanEval discordant conditions, the passing response scores about three points higher, with both intervals above zero. Apparent task success is intentionally a rubric dimension, so this is criterion-related evidence, not an independent human-preference validation. Functional execution remains authoritative for HumanEval.

\paragraph{Interpretation and scope.}
GroupEval separates judge-specific quality from local mode diversity. Multiple judges test sensitivity, not human alignment. Small groups can miss modes, and repeated judging does not replace independent generation. Human calibration of quality and validation of mode assignments remain open.

%% file: sections/app_theory.tex
\section{Finite-Step Analysis}
\label{app:theory}
The proofs follow \Cref{sec:theory}: \Cref{thm:accounting} in \Cref{app:accounting_proof}, \Cref{thm:generic} in \Cref{app:generic_proof}, and \Cref{prop:repair} in \Cref{app:compensation_proof}. Supporting bounds accompany the result they explain. \Cref{app:measurement_implementation} connects the theory to the posterior-loss measurement and implemented sampler.

\subsection{Preliminaries: assumptions and reveal paths}
\label{app:reveal_law}
\paragraph{Standing assumptions.}
All logarithms are natural. Entropy and KL use $0\log0=0$; conditional distributions on $P$-null events may be assigned arbitrarily. The vocabulary is finite, the sequence length $L$ is fixed unless stated otherwise, and denoisers assign positive probability to every clean token. Exact categorical arithmetic is assumed throughout the proofs; implementation perturbations are treated separately in \Cref{app:implementation}. The denoiser in \Cref{sec:theory} is written $q^i(\cdot\mid x_A)$; when a learned denoiser also receives the visible probability $a$, every statement below holds with $q^i(\cdot\mid x_A,a)$ in its place.

\paragraph{Reveal rounds and the augmented law.}
Let $0=a_0<a_1<\cdots<a_N=1$ and $\delta_k=a_k-a_{k-1}$. The ideal ancestral kernel reveals a still-masked coordinate in round $k$ with probability
\begin{equation}
 \rho_k=\frac{a_k-a_{k-1}}{1-a_{k-1}}.
 \label{eq:reveal_probability}
\end{equation}
Its survival probability through round $k$ is
$\prod_{j=1}^k(1-\rho_j)=1-a_k$, by telescoping. Hence the unconditional probability of first reveal in round $k$ is $\delta_k$. Independent reveal decisions across coordinates give independent $K_i\sim\mathrm{Cat}(\delta)$. The last round has $\rho_N=1$, so every coordinate is ultimately revealed.

For an assignment $K=(K_1,\ldots,K_L)$, write $S_k=\{i:K_i=k\}$ and $A_{k-1}=\cup_{j<k}S_j$. Empty blocks are allowed. Let $\nu(S)=\prod_i\delta_{K_i}$. Conditional on the complete partition, the sampler's clean-sequence law is
\begin{equation}
 Q_\lambda^S(x)=\prod_{k=1}^N\prod_{i\in S_k}
 q_\lambda^i(x_i\mid x_{A_{k-1}},a_{k-1}),\qquad
 Q_\lambda(x,S)=\nu(S)Q_\lambda^S(x).
 \label{eq:fixed_partition_law}
\end{equation}
Every factor is a normalized conditional distribution; summing in reverse block order shows $\sum_xQ_\lambda^S(x)=1$. The output law is $\pi_\lambda(x)=\sum_S\nu(S)Q_\lambda^S(x)$. Under the comparison law $P\otimes\nu$, the clean sequence and reveal partition are independent. Under the sampler they generally are not independent.

\subsection{Proof of \Cref{thm:accounting}: training integrates, sampling sums}
\label{app:accounting_proof}
We first establish the two area identities, then the finite-step decomposition. The final part derives the supporting bounds used in \Cref{sec:entropy_profile}.

\subsubsection{The entropy area identity and the training objective}
\label{app:profile_proof}
Revealing one more coordinate adds its conditional entropy. Averaging this observation over random visible sets gives the entropy profile as a derivative.

For any set function $f$ on subsets of $[L]$, let $U_a$ contain each coordinate independently with probability $a$. Its multilinear extension along the diagonal satisfies
\begin{equation}
 \frac{d}{da}\E f(U_a)
 =\sum_{i=1}^L\E_{A\subseteq[L]\setminus\{i\}}
 \big[f(A\cup\{i\})-f(A)\big],
 \label{eq:multilinear_derivative}
\end{equation}
where each element of $A$ is retained independently with probability $a$. To prove this, first allow distinct probabilities $a_1,\ldots,a_L$. Conditioning on inclusion of $i$ makes the expectation affine in $a_i$, with partial derivative equal to the bracketed difference. The chain rule along $a_1=\cdots=a_L=a$ gives \Cref{eq:multilinear_derivative}. This is an identity of finite polynomials, including at the endpoints.

Set $f(A)=H(X_A)$ and $F(a)=\E H(X_{U_a})$. Then
\begin{equation}
 F'(a)=\sum_i\E_A[H(X_{A\cup\{i\}})-H(X_A)]
 =\sum_i\E_AH(X_i\mid X_A)=h(a).
\end{equation}
Since $F(0)=0$ and $F(1)=H(X)$, integration proves
$\int_0^1h(a)\,da=H(X)$. The degree of $h$ is at most $L-1$.

At each visible context, cross-entropy splits into entropy and KL:
\begin{equation}
 e_\lambda(a)=\sum_i\E_{A,X_A}
 \KL{P(X_i\in\cdot\mid X_A)}{q_\lambda^i(\cdot\mid X_A,a)}\ge0.
 \label{eq:error_profile_full}
\end{equation}
This proves $c_\lambda=h+e_\lambda$ without an assumption that the denoiser is Bayes-optimal.

For the continuous-time connection, let the ideal masking schedule $\alpha_t$ decrease absolutely continuously from $\alpha_0=1$ to $\alpha_1=0$, and use the denoiser at the corresponding visible probability $a=\alpha_t$. At an interior time, the probability that a chosen coordinate is masked is $1-\alpha_t$. Conditional on its masking, the other coordinates remain independently visible with probability $\alpha_t$. Therefore the endpoint-completed negative ELBO functional is
\begin{align}
 \mathcal L_\infty(\lambda)
 &=\E_{X}\int_0^1\frac{-\alpha_t'}{1-\alpha_t}
 \E_{Z_t\mid X}\sum_i\mathbf1\{Z_t^i=m\}
 [-\log q_\lambda^i(X_i\mid Z_t,\alpha_t)]\,dt\notag\\
 &=\int_0^1(-\alpha_t')c_\lambda(\alpha_t)\,dt
 =\int_0^1c_\lambda(a)\,da.
 \label{eq:ct_change_variables}
\end{align}
The cancellation occurs before evaluating endpoint values. Measurability and integrability of $c_\lambda$ suffice; if the integral is infinite, the nonnegative identity still holds in the extended sense. In particular,
$\mathcal L_\infty(\lambda)-H(X)=\int_0^1 e_\lambda(a)\,da$.
This population functional does not assert that a clipped implementation logs the omitted endpoints.

\subsubsection{The finite-step decomposition}
\paragraph{Step 1: score a fixed reveal path.}
From \Cref{eq:fixed_partition_law}, cancellation of $\nu$ yields
\begin{align}
 \KL{P\otimes\nu}{Q_\lambda}
 &=\E_{S,X}\left[\log P(X)-\sum_k\sum_{i\in S_k}
 \log q_\lambda^i(X_i\mid X_{A_{k-1}},a_{k-1})\right].
 \label{eq:aug_expand}
\end{align}
Conditioned on $K_i=k$, the events $\{j\in A_{k-1}\}$ for $j\ne i$ are independent with probability $\sum_{r<k}\delta_r=a_{k-1}$. The conditional distribution of $A_{k-1}$ is consequently exactly the forced-mask context law in \Cref{eq:profiles}. Since $\Pr(K_i=k)=\delta_k$, the expected cross-entropy sum in \Cref{eq:aug_expand} equals $\sum_k\delta_k c_\lambda(a_{k-1})=\J_N(\lambda)$. Also $\E\log P(X)=-H(X)$. This proves \Cref{eq:augmented_identity} and, by $c_\lambda=h+e_\lambda$ and the area identity, its decomposition into $B_N+C_N(\lambda)$.

\paragraph{Step 2: identify the cost of same-round independence.}
To identify the parallelization term directly, define conditional total correlation by
\begin{equation}
 \TC(X_B\mid X_A)=\sum_{i\in B}H(X_i\mid X_A)-H(X_B\mid X_A).
 \label{eq:conditional_tc}
\end{equation}
It is the expected KL between the joint conditional law and the product of its marginals, hence is nonnegative. The empty-block value is zero. For every ordered partition,
$\sum_kH(X_{S_k}\mid X_{A_{k-1}})=H(X)$ by the entropy chain rule. Thus
\begin{align}
 \E_S\sum_k\TC(X_{S_k}\mid X_{A_{k-1}})
 &=\E_S\sum_k\sum_{i\in S_k}H(X_i\mid X_{A_{k-1}})-H(X)\notag\\
 &=\sum_k\delta_kh(a_{k-1})-H(X)=B_N.
\end{align}
Similarly, \Cref{eq:error_profile_full} identifies $C_N$ with the expected sum of per-token posterior KLs over revealed blocks. Both terms are nonnegative; only $C_N$ depends on the denoiser or temperature.

\paragraph{Step 3: pass from paths to outputs.}
Apply the KL chain rule in the opposite order, first conditioning on the output $X$:
\begin{equation}
 \KL{P\otimes\nu}{Q_\lambda}
 =\KL{P}{\pi_\lambda}
 +\E_{X\sim P}\KL{\nu}{Q_\lambda(S\mid X)}.
\end{equation}
Positive denoiser probabilities ensure $\pi_\lambda(x)>0$ and $Q_\lambda(S\mid x)>0$ for all relevant $x,S$. This proves \Cref{eq:marginal_gap}. In particular,
\begin{equation}
 \KL{P}{\pi_\lambda}\le B_N+C_N(\lambda),\qquad
 \TV(P,\pi_\lambda)\le\sqrt{\tfrac12[B_N+C_N(\lambda)]}.
 \label{eq:pinsker_certificate}
\end{equation}
These are upper bounds. Neither the temperature-independence of $B_N$ nor the convexity of $C_N$ transfers automatically to output KL, because the subtracted conditional KL $R_N$ depends on $\lambda$.

\subsubsection{Consequences for parallel sampling}
\label{app:dependence_profile}
\paragraph{Dependence profile and step-size bounds.}
Applying \Cref{eq:multilinear_derivative} to each summand of $h$ gives
\begin{align}
 D(a):=-h'(a)
 &=\sum_{i\ne j}\E_{A\subseteq[L]\setminus\{i,j\}}
 \left[H(X_i\mid X_A)-H(X_i\mid X_A,X_j)\right]\notag\\
 &=\sum_{i\ne j}\E_A I(X_i;X_j\mid X_A)\ge0.
 \label{eq:dependence_profile}
\end{align}
The sum is over ordered pairs. Each other coordinate is independently included in $A$ with probability $a$. This proves that $h$ is nonincreasing, without claiming that its derivative has a fixed monotonicity or that $h$ is convex.

On one grid interval $[b,d]$,
\begin{align}
 (d-b)h(b)-\int_b^dh(u)\,du
 &=\int_b^d[h(b)-h(u)]\,du
 =\int_b^d\int_b^uD(v)\,dv\,du\notag\\
 &=\int_b^d(d-v)D(v)\,dv.
\end{align}
Summing gives the exact refinement of the parallelization term
\begin{equation}
 B_N=\sum_{k=1}^N\int_{a_{k-1}}^{a_k}(a_k-u)D(u)\,du.
 \label{eq:exact_dependence_integral}
\end{equation}
Writing $\Delta=\max_k\delta_k$ and using $D\ge0$,
\begin{equation}
 B_N\le\Delta\int_0^1D(u)\,du
 =\Delta[h(0)-h(1)]
 =\Delta\sum_i I(X_i;X_{-i}).
 \label{eq:dependence_bound}
\end{equation}
The dependence sum equals total correlation plus dual total correlation:
\begin{equation}
 \sum_iI(X_i;X_{-i})=
 \underbrace{\sum_iH(X_i)-H(X)}_{\TC(X)}+
 \underbrace{H(X)-\sum_iH(X_i\mid X_{-i})}_{\mathrm{DTC}(X)}.
\end{equation}
This connects the Bernoulli-mask profile to the information quantities in prior schedule analyses~\citep{lavenant2025error,chen2025optimal}.

For an equal grid $a_k=k/N$, let
$T_N=\frac1N[\frac12h(0)+\sum_{k=1}^{N-1}h(k/N)+\frac12h(1)]$
be the composite trapezoidal rule. The left sum differs from $T_N$ by
$[h(0)-h(1)]/(2N)$. On any interval $[b,d]$, twice integrating by parts gives
\begin{equation}
 \frac{d-b}{2}[h(b)+h(d)]-\int_b^dh(u)\,du
 =\frac12\int_b^d(u-b)(d-u)h''(u)\,du.
\end{equation}
The absolute value is bounded by $\|h''\|_\infty(d-b)^3/12$. Consequently,
\begin{equation}
 \left|B_N-\frac{1}{2N}\sum_iI(X_i;X_{-i})\right|
 \le\frac{\|h''\|_\infty}{12N^2}.
 \label{eq:uniform_remainder}
\end{equation}
For fixed $P$ and $L$, this supplies the asymptotic coefficient, not only an $O(N^{-1})$ rate. The constant depends on $P,L$; this is not a dimension-uniform expansion when $L$ grows with $N$.

Inserting a grid point $b<c<d$ reduces its contribution to the left sum by $(d-c)[h(b)-h(c)]\ge0$. Thus nested refinement never increases $B_N$. This statement concerns the parallelization term, not the learned-denoiser term $C_N$ or the final output divergence. If $e_\lambda$ is $K$-Lipschitz, its quadrature has the separate standard bound
\begin{equation}
 \left|C_N(\lambda)-\int_0^1e_\lambda(a)\,da\right|
 \le\frac K2\sum_k\delta_k^2,
\end{equation}
obtained by integrating $|e_\lambda(a_{k-1})-e_\lambda(a)|\le K(a-a_{k-1})$ on each interval.

\paragraph{Entropy inflation for two tokens.}
\label{app:pair_entropy}
Let $L=2$, let $P$ be any law of $(X_1,X_2)$ with marginals $P_1,P_2$, and let $q$ be the Bayes denoiser at $\lambda=1$. The two coordinates share a round with probability $c=\sum_k\delta_k^2$, in which case each is drawn from its marginal. Otherwise one is revealed first from its marginal and the other from its exact conditional, which reproduces $P$ in either order. Hence $\pi_1=c\,P_1\otimes P_2+(1-c)P$. By concavity of entropy,
\begin{equation}
 H(\pi_1)\ge c\,H(P_1\otimes P_2)+(1-c)H(P)=H(P)+c\,I(X_1;X_2).
 \label{eq:pair_entropy}
\end{equation}
Parallel sampling therefore adds at least $c\,I(X_1;X_2)$ nats of spurious randomness to a perfect model. Entropy inflation does not by itself fix the improving temperature direction: a few binary two-token targets prefer slight smoothing, which is why \Cref{thm:generic} fixes no direction. We do not know whether \Cref{eq:pair_entropy} extends to $L\ge3$ under independent reveal rounds; $H(\pi_1)>H(P)$ held for all 280 random targets in \Cref{app:random_targets}.

\paragraph{Reveal-round collisions.}
\label{app:occupancy}
Under independent reveal rounds, the expected numbers of empty and singleton blocks, respectively, are
\begin{equation}
 \sum_k(1-\delta_k)^L,\qquad
 \sum_k L\delta_k(1-\delta_k)^{L-1}.
\end{equation}
A uniformly selected token shares its reveal round with another token with probability
$1-\sum_k\delta_k(1-\delta_k)^{L-1}$; the expected number of colliding unordered pairs is $\binom L2\sum_k\delta_k^2$. For $L=N=1024$ and equal masses, the expected empty and singleton counts are $376.53$ and $376.89$, the sharing probability is $.63194$, and the expected colliding-pair count is $511.5$. These describe sampling geometry, not a lower bound on output error or on actual model calls.

\subsection{Proof of \Cref{thm:generic}: the Bayes temperature is generically suboptimal}
\label{app:generic_proof}
\paragraph{Part (a).}
If every round reveals at most one coordinate, a reveal path is an ordering $\sigma$ of $[L]$, drawn independently of $X$. With Bayes conditionals, $Q_1^\sigma(x)=\prod_{j=1}^LP(x_{\sigma(j)}\mid x_{\sigma(1)},\ldots,x_{\sigma(j-1)})=P(x)$ by the chain rule, for every $\sigma$. Hence $\pi_1=P$ and $\lambda=1$ attains zero output KL. This includes left-to-right autoregressive decoding.

\paragraph{Part (b): one counterexample suffices.}
The derivative of output KL at $\lambda=1$ is analytic in the target law. Showing it is nonzero for just one target therefore proves that it is nonzero almost everywhere.
Parameterize full-support targets by the open simplex $\Delta^\circ\subset\mathbb R^{|\mathcal V|^L-1}$, a connected open set. Each Bayes conditional $P(x_i\mid x_A)=P(x_i,x_A)/P(x_A)$ is a ratio of linear functions with a positive denominator, hence real-analytic and positive on $\Delta^\circ$. Tempering, $q_\lambda(v)=\exp(\lambda\log q(v))/\sum_u\exp(\lambda\log q(u))$, is analytic in $(P,\lambda)\in\Delta^\circ\times(0,\infty)$. Each path law $Q_\lambda^S(x)$ is a finite product of such terms, $\pi_\lambda(x)=\sum_S\nu(S)Q_\lambda^S(x)$ is a positive finite sum, and $\KL{P}{\pi_\lambda}=\sum_xP(x)[\log P(x)-\log\pi_\lambda(x)]$ is therefore analytic. So is $g(P)=\partial_\lambda\KL{P}{\pi_\lambda}|_{\lambda=1}$. The zero set of a real-analytic function on a connected open set is either the whole set or Lebesgue-null~\citep{mityagin2015zero}. It is closed in $\Delta^\circ$ because $g$ is continuous. It remains to exhibit one target with $g\ne0$ for every $L\ge2$, $|\mathcal V|=V\ge2$, and grid with $N\ge2$.

\paragraph{A $V$-ary witness.}
Let $X_1$ be uniform on $\mathcal V$, let $X_2=X_1$ with probability $s\in(1/V,1)$ and otherwise be uniform on the other $V-1$ symbols, and let $X_3,\ldots,X_L$ be i.i.d.\ uniform and independent of $(X_1,X_2)$. This target has full support, and $X_2$ is marginally uniform. Bayes conditionals of $X_3,\ldots,X_L$ are uniform in every context and unchanged by tempering. The conditional of $X_1$ or $X_2$ is uniform when its partner is hidden, and otherwise the symmetric kernel $K(v\mid w)=s\,\mathbf1\{v=w\}+\frac{1-s}{V-1}\mathbf1\{v\ne w\}$, whose tempered version predicts agreement with probability
\begin{equation}
 u_\lambda=\frac{s^\lambda}{s^\lambda+(V-1)^{1-\lambda}(1-s)^\lambda}.
\end{equation}
The pair shares a round with probability $c=\sum_k\delta_k^2\in(0,1)$, since $N\ge2$ and every $\delta_k>0$. In that case both coordinates are uniform and independent; otherwise one is uniform and the other follows the tempered kernel, which is symmetric, so the order does not matter. Consequently $\pi_\lambda$ keeps $X_3,\ldots,X_L$ uniform and independent, keeps $X_1$ uniform, and makes the pair agree with probability $t_\lambda=c/V+(1-c)u_\lambda$, uniformly over agreeing and over disagreeing pairs. $P$ has the same structure with agreement probability $s$, so $\KL{P}{\pi_\lambda}=\KL{\Bern(s)}{\Bern(t_\lambda)}$ and
\begin{equation}
 g(P)=\frac{t_1-s}{t_1(1-t_1)}\,(1-c)\,u_1(1-u_1)\log\frac{s(V-1)}{1-s},\qquad t_1-s=c\Bigl(\frac1V-s\Bigr).
\end{equation}
For $s\in(1/V,1)$ the first factor is negative and the logarithm positive, so $g(P)<0$: sharpening strictly improves the Bayes sampler. For $V=2$ this is the family of \Cref{prop:repair}. This completes the proof.

\paragraph{An explicit form of the derivative.}
For a path $S$ and output $x$, let $A_i(S)$ be the coordinates revealed before the round of $i$, and define the path surprisal $\sigma(x,S)=-\log Q_1^S(x)=\sum_i-\log P(x_i\mid x_{A_i(S)})$, the path entropy $\eta(x,S)=\sum_iH(X_i\mid X_{A_i(S)}=x_{A_i(S)})$, and the path weight $w(x,S)=Q^S_1(x)/\pi_1(x)=Q_1(S\mid x)/\nu(S)$. Since $\partial_\lambda\log q_\lambda(v)|_{\lambda=1}=\log q(v)+H(q)$, we have $\partial_\lambda\log Q_\lambda^S(x)|_{\lambda=1}=\eta(x,S)-\sigma(x,S)$, and differentiating $\KL{P}{\pi_\lambda}=\sum_xP(x)\log P(x)-\sum_xP(x)\log\sum_S\nu(S)Q^S_\lambda(x)$ gives
\begin{equation}
 g(P)=\E_{P\otimes\nu}\bigl[w\,(\sigma-\eta)\bigr]
 =\operatorname{Cov}_{P\otimes\nu}(w,\sigma-\eta).
 \label{eq:derivative_covariance}
\end{equation}
The last equality uses $\E_\nu w(x,\cdot)=1$ for every $x$ and $\E_{P\otimes\nu}[\sigma-\eta]=0$, which holds because, under $P\otimes\nu$, each $X_i$ given $X_{A_i(S)}$ follows its Bayes conditional. For fixed $x$, $w=e^{-\sigma}/\pi_1(x)$ is decreasing in $\sigma$, so $\E_P\operatorname{Cov}_\nu(w,\sigma\mid X)\le0$ and
\begin{equation}
 g(P)=\E_P\operatorname{Cov}_\nu(w,\sigma\mid X)-\operatorname{Cov}_{P\otimes\nu}(w,\eta).
\end{equation}
The first term always pushes toward sharpening: the sampler over-weights the paths along which the data look least surprising. Reversing the direction requires a sufficiently negative covariance between path weight and conditional entropy: the favored paths must have lower, rather than higher, conditional entropy. $\lambda=1$ is stationary exactly when the two terms cancel.

\paragraph{Random-target enumeration.}
\label{app:random_targets}
The supplementary verifier enumerates every output and reveal assignment for $(L,|\mathcal V|,N)\in\{(2,2,2),(2,3,2),(2,2,3),(3,2,2),(3,2,3),(2,4,2),(3,3,2)\}$, 40 targets each, drawn from symmetric Dirichlet laws with concentration in $\{.3,1,3\}$, with random grids. For each target it checks the accounting identity of \Cref{thm:accounting}, compares \Cref{eq:derivative_covariance} with finite differences, confirms that $\lambda=1$ is not stationary, and locates the output-optimal $\lambda_\star$ on a logarithmic grid refined by bounded search. It also checks \Cref{eq:pair_entropy}, the $V$-ary witness in closed form, part (ii) of \Cref{prop:repair}, part (a) of \Cref{thm:generic}, and \Cref{prop:collapse} on random linear programs: 2,937 assertions in all, with maximum residual $4\times10^{-11}$. Sharpening is the improving direction for 279 of the 280 targets. The median $\lambda_\star$ is $1.16$ (5th--95th percentiles $1.003$--$1.69$), and the median relative reduction in output KL at $\lambda_\star$ is $5.6\%$ (95th percentile $39\%$). These are properties of small random targets, not estimates for language. 
% Code: \path{theory/verify_generic_suboptimality.py}.

\subsection{Proof of \Cref{prop:repair}: exact repair and inverted rankings}
\label{app:compensation_proof}
\paragraph{The two branches and their losses.}
For $P_s$ in \Cref{eq:binary_target}, both one-coordinate marginals are uniform. The Bayes probability that the second token matches the first, conditional on the first being visible, is $s$. Raising the two conditional probabilities to power $\lambda$ and normalizing yields
\begin{equation}
 u_\lambda=\frac{s^\lambda}{s^\lambda+(1-s)^\lambda}
 =\operatorname{sigmoid}(\lambda\logit s).
\end{equation}
With probability $c=\sum_k\delta_k^2$, both coordinates reveal in the same round and are drawn independently from uniform marginals. They then match with probability $1/2$. Otherwise, the first token is uniform and the second matches it with probability $u_\lambda$. Both branches are invariant under swapping coordinates and flipping both bits. Their mixture is therefore exactly $P_{t_\lambda}$ with $t_\lambda=c/2+(1-c)u_\lambda$. Summing the two diagonal and two off-diagonal KL contributions proves the first identity in \Cref{eq:binary_objectives}.

The unconditional marginals contribute no posterior error at any temperature. On a separate-round path, only the second token contributes error, namely
$\KL{\Bern(s)}{\Bern(u_\lambda)}$; this path has probability $1-c$. This proves the second identity. Since $s>1/2$, $u_\lambda$ is strictly increasing in $\lambda$ and equals $s$ exactly at $\lambda=1$. Hence $C_N$ is uniquely minimized at $1$ for $c<1$. The parallelization term is also explicit:
\begin{equation}
 B_N=c\,[\log2-\hbin(s)],\qquad
 \hbin(s)=-s\log s-(1-s)\log(1-s).
\end{equation}

\paragraph{(i) Exact repair.}
For the output objective, differentiation gives
\begin{equation}
 \frac{d}{d\lambda}\KL{\Bern(s)}{\Bern(t_\lambda)}
 =\frac{t_\lambda-s}{t_\lambda(1-t_\lambda)}
 (1-c)u_\lambda(1-u_\lambda)\logit s.
 \label{eq:binary_derivative}
\end{equation}
All factors other than $t_\lambda-s$ are positive. As $\lambda$ increases from $0$ to infinity, $t_\lambda$ strictly increases from $1/2$ to $1-c/2$ (the latter not attained at finite $\lambda$).

If $c<2(1-s)$, the target $s$ lies strictly between these endpoints. There is a unique zero of the derivative, at
$u_*=(s-c/2)/(1-c)$. It satisfies
$u_*-s=c(s-1/2)/(1-c)>0$, hence $\lambda_*>1$, and inverting the logistic map gives the expression in \Cref{prop:repair}(i). At that value $\pi_{\lambda_*}=P_s$, giving zero output KL and strictly positive $C_N(\lambda_*)$.

If $c\ge2(1-s)$, then $t_\lambda<s$ for every finite $\lambda$. The derivative is strictly negative, and the infimum occurs only at $\lambda\to\infty$. The limiting error is zero when $c=2(1-s)$ and strictly positive when $c>2(1-s)$. This proves both regimes. The excluded case $c=1$ reveals both coordinates together with probability one; temperature leaves the uniform product output unchanged and there is no strict posterior optimum.

\paragraph{(ii) Inverted ranking.}
In the finite-repair regime, consider any denoiser that, like the Bayes denoiser, predicts a uniform bit when the partner is hidden and predicts agreement with probability $u$ when the partner is visible. The argument above, with $u_\lambda$ replaced by $u$, gives output law $P_{t(u)}$ with $t(u)=c/2+(1-c)u$ and posterior loss $(1-c)\KL{\Bern(s)}{\Bern(u)}$, which is strictly positive for $u\ne s$. Because $t\mapsto\KL{\Bern(s)}{\Bern(t)}$ strictly decreases on $(0,s]$, and $t(s)<t(u)\le t(u_\star)=s$ for $u\in(s,u_\star]$, every such denoiser has strictly smaller output error than the Bayes denoiser at $\lambda=1$ and strictly larger posterior loss. Tempering either denoiser sweeps its agreement probability continuously over $(1/2,1)$, so both reach $u_\star$ and exactness at their own best temperatures; the inversion is a property of the shared fixed temperature.

\paragraph{(iii) Vanishing with steps.}
For fixed $s$, the finite-repair regime holds for all sufficiently small $c$. Taylor expansion of the explicit expression yields
\begin{equation}
 \lambda_*=1+
 \frac{s-1/2}{s(1-s)\logit s}\,c+O(c^2),\qquad c\to0.
 \label{eq:collision_expansion}
\end{equation}
For an equal $N$-round grid, $c=1/N$, so the output-optimal temperature tends to the Bayes value as collisions vanish. This is a property of the stated family, not a temperature schedule prescribed for language models.

\paragraph{Stability under perturbations.}
At any point in the finite-repair regime, fix the value $\lambda_*>1$. Its output KL is strictly smaller than at $\lambda=1$. On the interior of the finite probability simplex, Bayes conditionals, their tempered probabilities, the finite mixture output, and its KL are continuous functions of the target probabilities and grid masses. The strict improvement therefore persists in a neighborhood of that target and schedule. Exact zero-error repair relies on the symmetric family, but the existence of a sharpening benefit for a Bayes denoiser is stable to sufficiently small perturbations.

\subsection{Posterior-loss measurement and implementation}
\label{app:measurement_implementation}
We derive the estimator used in \Cref{sec:certificate}, then relate the ideal reveal grid to the implemented sampler.

\subsubsection{The forced-mask estimator and temperature derivatives}
\label{app:temperature_proof}
Draw $X\sim P$, $K\sim\mathrm{Cat}(\delta)$, and $I\sim\mathrm{Unif}([L])$, independently. Draw $A\subseteq[L]\setminus\{I\}$ by independent inclusion at probability $a_{K-1}$. The unbiased estimator is
\begin{equation}
 \widehat\J_N=L[-\log q_\lambda^I(X_I\mid X_A,a_{K-1})],
 \qquad \E\widehat\J_N=\J_N(\lambda).
 \label{eq:forced_mask_estimators}
\end{equation}
The factor $L$ cancels the uniform coordinate choice:
\begin{align}
 \E\{L[-\log q_\lambda^I(X_I\mid X_A,a_{K-1})]\}
 &=\sum_k\delta_k\sum_i\E_{A,X}[-\log q_\lambda^i(X_i\mid X_A,a_{k-1})]
 =\J_N(\lambda).
\end{align}
The target coordinate is always masked by construction. No inverse factor $1/(1-a)$ is needed. This establishes unbiasedness, not a universal variance comparison with every alternative estimator.

Temperature scaling is convex in inverse temperature for fixed logits~\citep{guo2017calibration}; here those contexts are weighted by the reveal grid. For a fixed context and target $x$, let $r(v)$ be finite logits and
$q_\lambda(v)=\exp(\lambda r(v))/\sum_u\exp(\lambda r(u))$. Then
\begin{align}
 \frac{d}{d\lambda}[-\log q_\lambda(x)]
 &=\E_{Y\sim q_\lambda}r(Y)-r(x),\label{eq:temp_first}\\
 \frac{d^2}{d\lambda^2}[-\log q_\lambda(x)]
 &=\Var_{Y\sim q_\lambda}r(Y)\ge0.\label{eq:temp_second}
\end{align}
At a finite grid there are finitely many contexts, so differentiation commutes with their weighted sum. Therefore
\begin{equation}
 \J_N'(\lambda)=L\E[\E_{Y\sim q_\lambda^I}r^I(Y)-r^I(X_I)],
 \qquad
 \J_N''(\lambda)=L\E\Var_{Y\sim q_\lambda^I}r^I(Y).
\end{equation}
The first expression is also an unbiased forced-mask gradient estimator. If at least one positively weighted context has nonconstant logits, its variance is strictly positive at every finite $\lambda$, because all token probabilities are positive. Then $\J_N$ is strictly convex. A minimizer exists on every nonempty compact inverse-temperature interval and is unique under this strictness condition. An unconstrained finite minimizer on $(0,\infty)$ is not guaranteed.

For an exact Bayes denoiser, $C_N(1)=0$, so $\lambda=1$ minimizes $\J_N$. It is unique when a positively weighted Bayes conditional is nonuniform (and positive). Conversely, a negative derivative $\J_N'(1)$ supports sufficiently small increases in inverse temperature, not an arbitrary finite sharpening. A positive measured finite difference $\J_N(1/.9)-\J_N(1)$ does not, by itself, determine the derivative at $1$. Our empirical statement is the finite contrast, not a claim to exclude every possible infinitesimal calibration adjustment.

\subsubsection{Schedules, precision, and verification}
\label{app:implementation}
\paragraph{Configured updates and endpoint completion.}
The empirical MDLM code uses a clipped linear schedule, starts from all masks, and applies $M$ updates followed by terminal cleanup at $\epsilon=10^{-5}$. The code uses $\alpha(t)=1-(1-\eta)t$, where $\eta=10^{-3}$ is the schedule clipping constant, distinct from the terminal time $\epsilon$. For its idealized categorical counterpart, a masked token survives an update $t\to s$ with probability $(1-\alpha(s))/(1-\alpha(t))=s/t$. Starting from all masks at $t=1$, its survival probability at time $t$ is therefore $t$. The corresponding visible probability is $a=1-t$, rather than $\alpha(t)$. A uniform time grid from $1$ to $\epsilon$ consequently gives $M$ ordinary reveal masses $(1-\epsilon)/M$ and terminal mass $\epsilon$, so $N=M+1$. The time-independent MDLM predictor allows this reparameterization without changing its logits. The proper-loss measurement uses these context weights; it does not measure the fp32 output kernel's KL. Probability floors, roundoff, and categorical implementation effects are not included in the exact law above.

If two procedures have \emph{identical} laws before terminal cleanup and differ only in how they fill surviving masks, coupling their common prefix gives output-TV difference at most
$\Pr(\text{some mask survives})=1-(1-\epsilon)^L\le L\epsilon$.
For $L=1024$ this is approximately $.01019$. This comparison does not justify replacing a whole earlier grid by a different grid: their prefix laws need not agree.

\paragraph{Categorical precision.}
The implementation uses one categorical draw over clean tokens and the mask outcome. Finite precision can therefore change both token probabilities and reveal timing; the independent reveal law $\nu$ above describes the exact-arithmetic counterpart. The following comparison treats these changes jointly as perturbations of the complete transition kernel.
Suppose exact and implemented transition kernels $P_k$ and $\widetilde P_k$ obey
$\sup_z\TV(P_k(z,\cdot),\widetilde P_k(z,\cdot))\le\xi_k$ with $\xi_k\in[0,1]$ at every step, including cleanup. A stepwise maximal coupling, conditional on no previous disagreement, succeeds at step $k$ with probability at least $1-\xi_k$. Thus
\begin{equation}
 \TV(\pi,\widetilde\pi)\le1-\prod_k(1-\xi_k)\le\sum_k\xi_k.
\end{equation}
No numerical value of $\xi_k$ is assumed or estimated here. Finite arithmetic may alter support, so a TV perturbation statement cannot be substituted for a finite forward-KL guarantee. In particular, fp32 is an experimental operating point, not a mathematically exact implementation of a universal lower temperature.

\paragraph{Verification artifacts.}
The supplementary verifiers enumerate tiny binary and ternary distributions, including distributions with zero-probability sequences, and nonuniform reveal grids. They check augmented and marginal KL identities, conditional total correlation, convex-temperature derivatives, finite-grid bounds, both regimes and the equality boundary of the collision example, stability under asymmetric target perturbations, the clipped-schedule reveal masses, and the random-target results of \Cref{app:random_targets}. Separate integer/Fraction coefficient checks verify the entropy-area and conditional-mutual-information identities for formal subset entropies through $L=8$. These checks complement the proofs above; they are not proof-assistant certification.

%% file: sections/app_samplers.tex
\section{Extended Related Work and Implications for Sampler Design}
\label{app:samplers}\label{sec:related}
Our findings connect three research questions. How much of the few-step gap requires a new model? Which sampling changes can close it? And which measurements can tell whether it has closed? We develop these connections below, extending the recommendations in \Cref{sec:discussion}.

\subsection{Stronger baselines and the few-step gap}
Few-step quality of masked diffusion~\citep{austin2021d3pm,campbell2022continuous,sahoo2024simple,shi2024simplified,ou2024radd} has been pursued through distillation~\citep{deschenaux2024sdtt,hayakawa2025di4c,sahoo2025duo,zheng2025didi}, consistency training~\citep{amin2026consistent}, and alternative processes~\citep{lou2024sedd,vonrutte2025gidd,pynadath2025candi}. These methods change the trained generator. Our audit asks how much the original checkpoint can already do when its sampler is tuned.

For language, \citet{zheng2024masked} identified fp32 truncation and advocated fp64. CANDI showed that temperature can trade for steps along perplexity--entropy frontiers~\citep{pynadath2025candi}, and CaRE found that temperature explains most MAUVE variance among remasking strategies~\citep{shah2026care}. Our results establish that the tuned baseline rivals its distilled successors, that the gain survives semantic evaluation, and that a departure from temperature one helps even exact posteriors under parallel sampling. The public MDLM moves from gPPL $107.1$ at 1,024 standard steps to $41.3$ at 64 sharpened steps (\Cref{tab:headline_main}). At 256 steps it also exceeds the standard sampler at 1,024 on \textsc{GEQ} under every judge (\Cref{sec:ge_results}). The baseline correction thus survives a change of measurement.

Related evidence comes from exact test-time likelihoods, which find MDMs better than previously thought~\citep{turok2026duel}, and from analyses tracing much of the masked--uniform gap to parameterization and sampling design~\citep{gourevitch2026uniform}. Tuning baselines has closed apparent gaps before, for GANs~\citep{lucic2018gans} and neural language models~\citep{melis2018state}. Here the correction also exposes an inversion between predictor accuracy and generation quality.

\subsection{Sampler design and the price of parallelism}
A growing literature improves pretrained diffusion LMs at inference alone. Theory bounds factorization error through information profiles and total correlation~\citep{lavenant2025error,wainwright2026geometry}, characterizes it exactly for any schedule~\citep{chen2025optimal}, relates it to generation order~\citep{zhang2026generation}, and shows that it can escape likelihood-based metrics~\citep{tang2026sampler}. \Cref{thm:accounting} extends the accounting to learned, tempered denoisers and isolates the temperature-dependent credit for mixing paths. This separates improving posterior calibration~\citep{platt1999probabilistic,guo2017calibration} from improving the generated distribution.

\paragraph{One accounting, several ways to improve the sampler.}
For non-adaptive reveal rounds, output error is $B_N+C_N(\lambda)-R_N(\lambda)$. This decomposition distinguishes the mechanisms that sampler design can exploit.
\begin{itemize}
\setlength\itemsep{1pt}
\item \textbf{Schedule.} $B_N$ depends only on the data and grid. Its mesh bound~\eqref{eq:dependence_bound} and the fact that nested refinement never increases it (\Cref{app:dependence_profile}) explain why schedule design~\citep{lavenant2025error,chen2025optimal,wainwright2026geometry} helps.
\item \textbf{Temperature and truncation.} Exact temperature scaling leaves the reveal law and $B_N$ unchanged and trades posterior error $C_N$ against path credit $R_N$ (\Cref{prop:repair}). Truncation can also sharpen predictions. Finite-precision categorical draws can additionally change reveal timing, so fp32 is not equivalent to a scalar temperature adjustment (\Cref{app:implementation}).
\item \textbf{Order and dependence-aware selection.} Samplers plan or adapt the unmasking order~\citep{liu2025ddpd,kim2025train,benhamu2025eb,hayakawa2026maskgit}, or space and screen reveal sets for dependence~\citep{luxembourg2026dus,azangulov2025punt,sahin2026adas,kim2025klass}. They target the same-round dependence that $B_N$ prices. State-dependent reveal laws fall outside \Cref{thm:accounting} as stated.
\item \textbf{Joint draws.} Learned joint samplers~\citep{bansal2025joint} and mixture kernels~\citep{ahmed2026lkf} model dependence among same-round tokens, addressing the factorization that gives rise to $B_N$.
\item \textbf{Revision.} Remasking and predictor--corrector samplers~\citep{wang2025remdm,deschenaux2026psi} revisit committed tokens, which also lies outside our accounting.
\end{itemize}
These approaches can complement sharpening. Their gains should be measured after controlling the baseline's sharpness, so that a new mechanism receives credit for what it adds.

\paragraph{Temperature confounds sampler comparisons.}
Sharpening alone substantially improves the baseline, and temperature one is generically suboptimal even for an exact denoiser under parallel sampling (\Cref{thm:generic}). Some samplers also change sharpness implicitly. Confidence-based selection carries an implicit temperature~\citep{hayakawa2026maskgit}, and fp32 draws truncate categorical noise~\citep{zheng2024masked}. A comparison against a baseline at $\tau=1$ can therefore credit a new sampler with a temperature gain. Four practices help separate the effects.
\begin{enumerate}
\setlength\itemsep{1pt}
\item Tune temperature and precision for both the baseline and proposed sampler under the same screen, disclose the search budget, and report the gain over the baseline's best setting.
\item Report configured steps and network calls separately, together with measured latency and memory. Caching and adaptive reveal can change their relationship (\Cref{app:setup}).
\item Supplement gPPL and unigram entropy with group-level evaluation. \Cref{prop:collapse} shows why matching token entropy does not rule out collapse.
\item Evaluate checkpoints together with their samplers. Held-out posterior loss sees $B_N+C_N$ but not $R_N$ (\Cref{sec:certificate}). Exact test-time likelihood, where available~\citep{turok2026duel}, instead scores the output law itself.
\end{enumerate}
For the tested 8B diffusion LMs decoded greedily, positive temperatures do not help (\Cref{app:large_models}). The temperature comparison is most relevant to stochastic sampling. Under greedy decoding, order, selection, and revision remain available, and our analysis does not attribute a particular adaptive sampler's gain to temperature.

\subsection{Evaluating quality and diversity across outputs}
Quality--diversity trade-offs and the limits of surface metrics are well documented~\citep{hashimoto2019huse,holtzman2020nucleus,tevet2021diversity}. Reference-based~\citep{pillutla2021mauve}, kernel~\citep{friedman2023vendi}, meaning-cluster~\citep{kuhn2023semantic}, and distinct-response~\citep{zhang2025noveltybench} measures capture related notions. LLM judges provide scalable but biased ratings~\citep{liu2023geval,zheng2023judging,wang2024fair}. GroupEval combines blinded group-level mode partitioning, validity-weighted entropy, and a separate quality score for unconditional generation, where no prompt defines equivalence. Its contribution is a fixed, reusable protocol and the research conclusions it changes.

Likelihood is known to favor repetitive text~\citep{holtzman2020nucleus}. The DiDi comparison shows that token entropy need not detect this failure, even when two likelihood evaluators agree. GroupEval also distinguishes between sharpening gains that survive semantic judgment and those that do not. Its quality and diversity scores remain separate because good individual outputs and a varied output distribution are distinct requirements. Human calibration and validation of the mode assignments remain open, while the released rubric, grouping rules, and scorer make this distinction directly testable under a common protocol.